\documentclass[10pt,twocolumn,letterpaper]{article}

\usepackage{cvpr}              

\usepackage[accsupp]{axessibility} 

\usepackage{multirow}
\usepackage{pifont}

\usepackage[dvipsnames]{xcolor}

\definecolor{cvprblue}{rgb}{0.21,0.49,0.74}
\usepackage[pagebackref,breaklinks,colorlinks,citecolor=cvprblue]{hyperref}

\def\paperID{6467} 
\def\confName{CVPR}
\def\confYear{2024}

\title{Tri-Perspective View Decomposition for Geometry-Aware Depth Completion}

\author{Zhiqiang Yan$^{1}$, Yuankai Lin$^{2}$, Kun Wang$^{1}$, Yupeng Zheng$^{3}$, Yufei Wang$^{4}$, \\ Zhenyu Zhang$^{5}$, Jun Li$^{1}$\thanks{Corresponding authors}, and Jian Yang$^{1}$\footnotemark[1]\\
$^1$PCA Lab\thanks{PCA Lab, Key Lab of Intelligent Perception and Systems for High-Dimensional Information of Ministry of Education, and Jiangsu Key Lab of Image and Video Understanding for Social Security, School of Computer Science and Engineering, Nanjing University of Sci. \& Tech.}\ , Nanjing University of Science and Technology, China\\
$^2$Huazhong University of Science and Technology \ \ \ $^3$Chinese Academy of Sciences\\
$^4$Northwestern Polytechnical University \ \ \ $^5$Nanjing University\\
{\tt\small \{yanzq, kunwang, junli, csjyang\}@njust.edu.cn, zhangjesse@foxmail.com}\\
{\tt\small linyuankai@hust.edu.cn, zhengyupeng2022@ia.ac.cn, wangyufei1951@gmail.com}\\
}

\begin{document}
\maketitle


\begin{abstract}

Depth completion is a vital task for autonomous driving, as it involves reconstructing the precise 3D geometry of a scene from sparse and noisy depth measurements. However, most existing methods either rely only on 2D depth representations or directly incorporate raw 3D point clouds for compensation, which are still insufficient to capture the fine-grained 3D geometry of the scene. To address this challenge, we introduce \textbf{T}ri-\textbf{P}erspective \textbf{V}iew \textbf{D}ecomposition (\textbf{TPVD}), a novel framework that can explicitly model 3D geometry. In particular, (1) TPVD ingeniously decomposes the original point cloud into three 2D views, one of which corresponds to the sparse depth input. (2) We design TPV Fusion to update the 2D TPV features through recurrent 2D-3D-2D aggregation, where a Distance-Aware Spherical Convolution (DASC) is applied. (3) By adaptively choosing TPV affinitive neighbors, the newly proposed Geometric Spatial Propagation Network (GSPN) further improves the geometric consistency. As a result, our TPVD outperforms existing methods on KITTI, NYUv2, and SUN RGBD. Furthermore, we build a novel depth completion dataset named TOFDC, which is acquired by the time-of-flight (TOF) sensor and the color camera on smartphones. \href{https://yanzq95.github.io/projectpage/TOFDC/index.html}{Project page}.

\end{abstract}    
\section{Introduction}\label{sec:introduction}
Depth completion \cite{Uhrig2017THREEDV}, the technique of recovering dense depth maps from sparse ones, has a variety of applications in computer vision, such as scene understanding \cite{2018Sparse,Xu2019Depth,yan2022learning,liu2023single,zhao2023spherical,shao2023nddepth}, 3D reconstruction \cite{ma2018self,yan2022multi,rho2022guideformer,zhao2022discrete,yan2023distortion}, and autonomous driving \cite{wang2021regularizing,yan2022rignet,zhang2023cf,zheng2023steps,yan2023desnet,yan2023learnable,zhou2023bev}. All of these applications are highly dependent on accurate and reliable depth predictions. However, due to the constraints of hardware development and challenging environment, depth sensors are unable to deliver pixel-wise depth feedback, particularly in outdoor scenarios where the depth density is as low as 5\%. Thus, it is essential and worthwhile to complete the void areas of sparse depth for realistic applications. 

\begin{figure}[t]
 \centering
 \includegraphics[width=0.92\columnwidth]{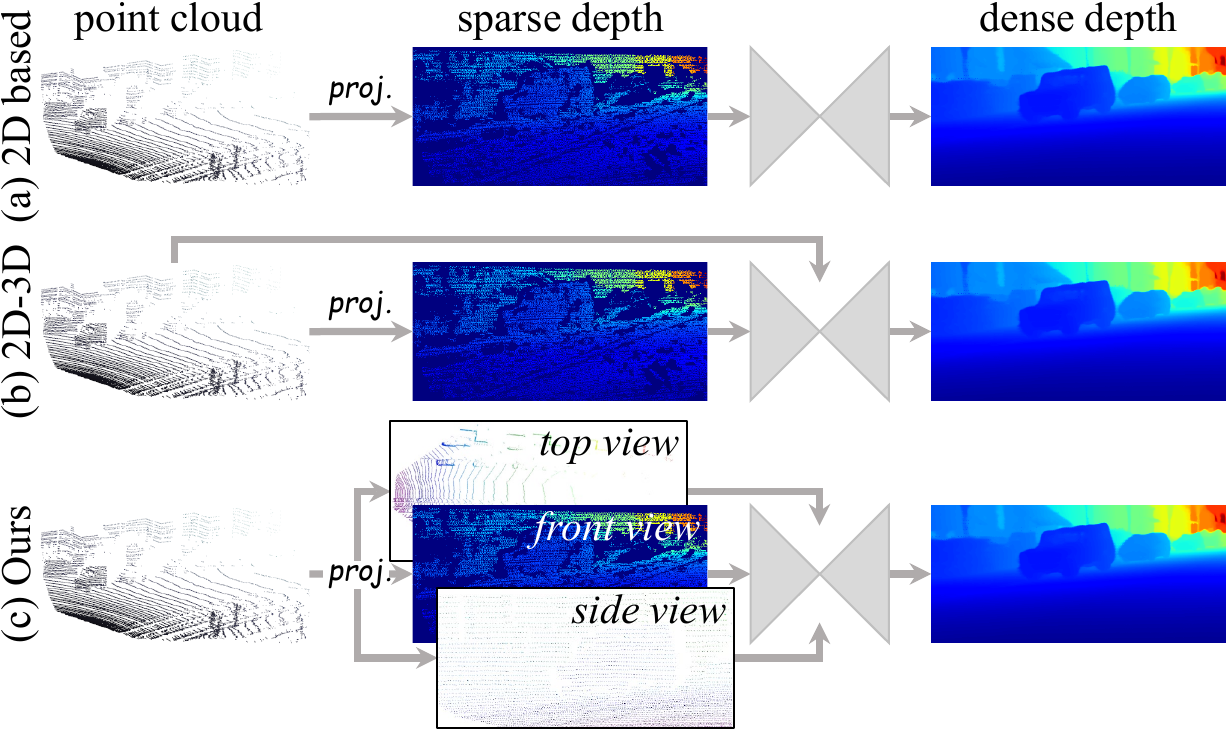}\\
 \caption{Framework comparison. (a) Previous 2D methods focus on 2D space to recover dense depth, and (b) recent 2D-3D joint approaches introduce 3D point clouds for assistance. Differently, (c) our TPVD decomposes the 3D point clouds into three 2D views to densify the sparse input while preserving the 3D geometry.}\label{framework_comparison}
\end{figure}

Most previous depth completion methods \cite{ma2018self,Cheng2020CSPN++,park2020nonlocal,hu2020PENet,zhang2023cf,yan2023rignet++,chen2023agg,wang2023lrru} focus on 2D feature space to learn depth representations, leading to a severe lack of 3D geometric information. As an alternative, some recent approaches \cite{chen2019learning,Qiu_2019_CVPR,zhao2021adaptive,huynh2021boosting,liu2022graphcspn,zhou2023bev,yu2023aggregating} attempt to incorporate 3D geometric priors directly from raw point clouds, rather than relying only on 2D representations. For example, \cite{zhou2023bev,yu2023aggregating} extract point cloud features to embed 3D geometry into their 2D depth generation branches. However, as we known that the point clouds are extremely sparse and their point distributions are varying in different distances, both of which deeply impede the performance of recent models. 

To address the above issues, we propose a novel framework called tri-perspective view decomposition (TPVD). As shown in Fig.~\ref{framework_comparison}, unlike existing 2D-3D joint methods \cite{chen2019learning,zhou2023bev,yu2023aggregating}, TPVD cleverly decomposes the 3D point clouds into three 2D views: top, front, and side. It is worth mentioning that the sparse depth input is exactly included in the front-view map. This decomposition enables TPVD to densify the sparse 3D point clouds in 2D space using 2D convolutions. To leverage the 3D geometric priors more effectively, TPVD employs a recurrent 2D-3D-2D TPV Fusion scheme. In this scheme, the denser 2D TPV features are projected back to 3D space to obtain coarse structural representations. Then, a distance-aware spherical convolution (DASC) is applied to encode the points with varying distributions in a compact spherical space, contributing to refined geometric structures. Next, the 3D spherical features are re-projected into 2D space to update the initial 2D TPV features. That is to say, the 2D process predicts more valid pixels to enrich the 3D process with denser points, while the 3D process captures geometry and feeds it back to the 2D process. These two processes complement each other. 

Moreover, TPVD incorporates a plug-and-play geometric spatial propagation network (GSPN) for full-scale 3D geometric refinement. Unlike previous 2D SPN \cite{xu2020deformable,Cheng2020CSPN++,park2020nonlocal,lin2023dyspn} and 3D SPN \cite{liu2022graphcspn,zhou2023bev} methods that generate their affinitive neighbors in either a single 2D space or a bird’s-eye view space, GSPN constructs the affinity simultaneously in the three decomposed 2D TPV spaces and their joint 3D projection space. Therefore, the affinity preserves both the neighborhood information and the 3D geometric structures. 

In addition, since depth information plays a crucial role in accurate 3D reconstruction and human-computer interaction, time-of-flight (TOF) depth sensors are increasingly equipped on edge mobile devices. In this paper, we collect a new depth completion dataset termed TOFDC, with a smartphone that has both TOF lens and color camera. 

In summary, our contributions are as follows:
\begin{itemize}
    \item We introduce a novel framework termed \textit{TPVD}, which densifies the sparse input whilst retaining 3D geometry. 
    \item \textit{TPV Fusion} is proposed to leverage the 3D geometry effectively via recurrent 2D-3D-2D interaction, where \textit{DASC} is applied to handle the varying distributions of LiDAR points. Besides, we design \textit{GSPN} to further produce fine-grained 3D geometric structures. 
    \item We build \textit{TOFDC}, a new smartphone-based depth completion dataset. Moreover, our method consistently outperforms the state-of-the-art approaches on four datasets: KITTI, NYUv2, SUN RGBD, and TOFDC. 
\end{itemize}

\section{Related Work}\label{sec:related_work}

\noindent\textbf{2D Based Depth Completion.} 
Usually, the sparse depth is taken from structured light \cite{silberman2012indoor}, TOF \cite{he2021towards},  LiDAR \cite{Uhrig2017THREEDV,caesar2020nuscenes}, stereo cameras \cite{geiger2012we,mayer2016large}, or structure from motion \cite{wong2020unsupervised,schonberger2016structure}. 

Recent 2D based image-guided methods~\cite{2020Confidence,hu2020PENet,imran2021depth,wang2023lrru} focus on RGB-D fusion by direct concatenation or summation. Differently, GuideNet~\cite{tang2020learning} adopts a guided filtering, whose kernel weight is from the guided RGB image.
FCFRNet \cite{liu2021fcfr} designs an energy-based fusion to integrate the RGB-D features. RigNet~\cite{yan2022rignet} and RigNet++~\cite{yan2023rignet++} propose a new guidance unit with low complexity to produce the dynamic kernel. 
GFormer~\cite{rho2022guideformer} and CFormer~\cite{zhang2023cf} concurrently leverage convolution and transformer to extract both local and long-range representations. Most recently, LRRU~\cite{wang2023lrru} presents a large-to-small dynamical kernel scope to capture long-to-short dependencies. However, these 2D based methods deployed in 2D space cannot reserve very precise 3D spatial geometric information.

\noindent\textbf{2D-3D Joint Depth Completion.} 
It is more intuitive and effective to capture geometric structures with 3D representations, such as surface normals~\cite{Xu2019Depth,Qiu_2019_CVPR}, graphs~\cite{zhao2021adaptive,liu2022graphcspn}, point clouds~\cite{chen2019learning,huynh2021boosting,yu2023aggregating}, and voxels~\cite{zhou2023bev}. 

For the first time, DLiDAR~\cite{Qiu_2019_CVPR} and DepthNormal~\cite{Xu2019Depth} introduce surface normals to boost the performance. In view of the effectiveness of the graph neural networks in representing neighborhood relation, ACMNet~\cite{zhao2021adaptive} applies attention-based graph propagation for multi-modal fusion. GraphCSPN~\cite{liu2022graphcspn} leverages convolution neural networks as well as graph neural networks in a complementary way for geometric learning. Lately, FuseNet~\cite{chen2019learning} and PointDC~\cite{yu2023aggregating} involve LiDAR point cloud branches to model 3D geometry. Moreover, BEV@DC~\cite{zhou2023bev} adopts point-voxel architecture based on bird's-eye view for better effectiveness-efficiency trade-off. Different from these 2D-3D joint methods, our TPVD restores dense 2D depth in 2D space while retaining the 3D geometric priors via point cloud decomposition.

\begin{figure*}[t]
 \centering
 \includegraphics[width=2.08\columnwidth]{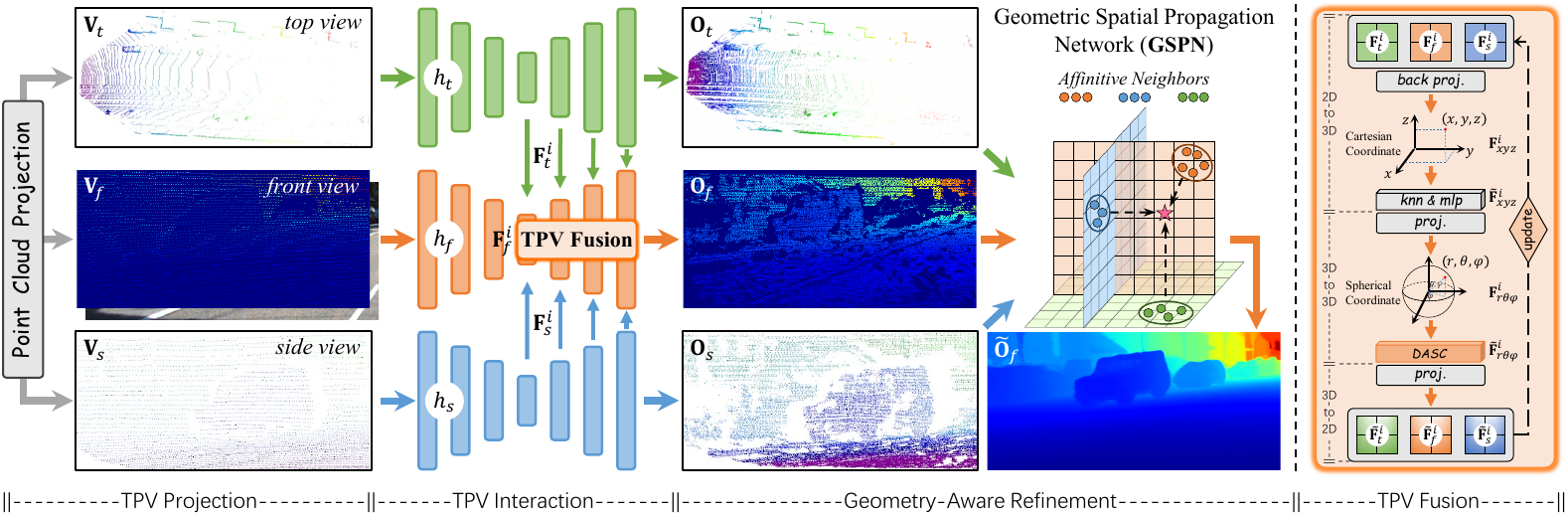}\\
 \caption{Pipeline of TPVD. The 3D point cloud is first projected into top, side, and front views, where the raw 2D sparse depth input is included in the front view. Then the three views are fed into 2D UNets to produce TPV features that are aggregated by the 2D-3D-2D TPV Fusion, obtaining denser depth with richer geometry. Finally, on the output side, the plug-and-play geometric spatial propagation network (GSPN) generates refined depth results with consistent geometry. \textit{DASC} refers to the distance-aware spherical convolution.}\label{pipeline}
\end{figure*}

\noindent\textbf{Spatial Propagation Network.} 
SPN~\cite{Cheng2020CSPN++} is increasingly emerging in both 2D based~\cite{park2020nonlocal,lin2023dyspn,yan2023rignet++} and 2D-3D joint~\cite{liu2022graphcspn,zhou2023bev} depth completion methods. It digs local or non-local neighbors by 2D and 3D anisotropic filtering kernels. 

Initially, 2D SPNs~\cite{liu2017SPN} are first proposed to learn pairwise similarity matrix. CSPN~\cite{2018Learning} conducts recursive convolutions with fixed local neighborhood kernels for improvement, while CSPN++~\cite{Cheng2020CSPN++} learns adaptive kernel sizes. PENet~\cite{hu2020PENet} further enlarges the receptive fields with dilated convolutions. Differently, NLSPN~\cite{park2020nonlocal} incorporates non-local neighbors via deformable convolutions. Similarly, DySPN~\cite{lin2023dyspn} produces dynamic non-linear neighbors by attention mechanism. 3D SPNs~\cite{cspn_pami} are commonly embedded in 2D-3D joint methods to utilize 3D geometry. For example, S3CNet~\cite{cheng2021S3CNet} computes key spatial features from LiDAR by a 3D spatial propagation unit. GraphCSPN~\cite{liu2022graphcspn} uses geometric constraints to regularize the 3D propagation. Recently, BEV@DC~\cite{zhou2023bev} conducts a point-voxel spatial propagation network for 3D dense supervision. Differently, we aggregate the 2D affinitive neighbors in 2D TPV spaces, resulting in gradual refinement of 3D geometric awareness.

\section{TPVD}
\subsection{Overview}
Recent works \cite{chen2019learning,zhou2023bev,yu2023aggregating} tend to introduce 3D point clouds to boost the 2D depth completion. Differently, this paper restores dense depth mainly in 2D space, whilst retaining the 3D geometric priors via point cloud decomposition. 

Fig.~\ref{pipeline} shows our pipeline that consists of \ding{192} TPV projection, \ding{193} TPV interaction, and \ding{194} geometry-aware refinement. Specifically, in \ding{192} the 3D point cloud is first projected into top, side, and front sparse depth views. Then in \ding{193} three subnetworks are employed to extract the TPV features, where the TPV Fusion with a distance-aware spherical convolution (DASC) is designed to leverage the 3D geometric priors. Finally, to obtain dense completion with more fine-grained geometry, in \ding{194} the geometric spatial propagation network (GSPN) further improves the geometric consistency.

\subsection{TPV Projection}
Given a 2D sparse depth map $\mathbf{S}\in {\mathbb{R}}^{H\times W}$ with the binary mask $m$, we first transform it into a 3D point cloud, which is then processed by a Multi-layer Perceptron (MLP) and two continuous convolutions \cite{chen2019learning} to generate the point feature $\mathbf{P}\in {\mathbb{R}}^{N \times 3}$. Then we employ $\mathcal{P}_{tpv}$ to project the 3D $\mathbf{P}$ into 2D orthogonal top-view $\mathbf{V}_t\in {\mathbb{R}}^{W \times D}$, side-view $\mathbf{V}_s\in {\mathbb{R}}^{D \times H}$, and front-view $\mathbf{V}_f\in {\mathbb{R}}^{H \times W}$. Particularly, we combine $\mathbf{S}$ and $\mathbf{V}_f$ via the mask $m$ to update $\mathbf{V}_f$:
\begin{equation}\label{eq_tpv_projection}
\begin{split}
    &\mathbf{V}_t, \ \mathbf{V}_s, \ \mathbf{V}_f=\mathcal{P}_{tpv}(\mathbf{P}),\\
    &\text{new} \ \mathbf{V}_f=\mathbf{S}+(1-m)\mathbf{V}_f.
\end{split}
\end{equation}
Unless stated, we use $\mathbf{V}_f$ to represent the new $\mathbf{V}_f$ below. 

\subsection{TPV Interaction}
In Fig.~\ref{pipeline}, we use $h_t$, $h_s$, and $h_f$ subnetworks to encode $\mathbf{V}_t$, $\mathbf{V}_s$, and $\mathbf{V}_f$, as well as the image $\mathbf{I}$ that is aligned with $\mathbf{V}_f$. In each $i$th layer of the three decoders, their intermediate features are severally denoted as $\mathbf{F}_t^i \in {\mathbb{R}}^{W_i \times D_i \times C_i}$, $\mathbf{F}_s^i \in {\mathbb{R}}^{D_i \times H_i \times C_i}$, and $\mathbf{F}_f^i \in {\mathbb{R}}^{H_i \times W_i \times C_i}$. While $1\leq i\leq 4$: 
\begin{equation}\label{eq_tpv_decoder}
\mathbf{F}_t^i, \ \mathbf{F}_s^i, \ \mathbf{F}_f^i=h_t(\mathbf{V}_t), \ h_s(\mathbf{V}_s), \ h_f(\mathbf{V}_f, \ \mathbf{I}).
\end{equation}

\noindent\textbf{TPV Fusion.} After obtaining the three 2D TPV features, we introduce TPV Fusion. In Fig.~\ref{pipeline} (right), there are three steps in a single iteration of the fusion process:

(1) \textit{2D-to-3D}: To learn 3D geometric priors, the 2D $\mathbf{F}_t^i$, $\mathbf{F}_s^i$, and $\mathbf{F}_f^i$ are jointly projected back to the 3D Cartesian coordinate, yielding $\mathbf{F}_{xyz}^i$. Then, the k-Nearest Neighbor (KNN) computes the $k$ relevant neighbors, while MLP further maps the aggregated features, obtaining the 3D $\tilde{\mathbf{F}}_{xyz}^i$: 
\begin{align}
    &\mathbf{F}_{xyz}^i=\mathcal{P}_{tpv}^{-1}(\mathbf{F}_t^i, \ \mathbf{F}_s^i, \ \mathbf{F}_f^i),\label{eq_step1-1}\\
    &\tilde{\mathbf{F}}_{xyz}^i=h_{km}(\mathbf{F}_{xyz}^i),\label{eq_step1-2}
\end{align}
where $h_{km}\left( \cdot \right)$ denotes the combined KNN and MLP. 

\begin{figure}[t]
 \centering
 \includegraphics[width=0.95\columnwidth]{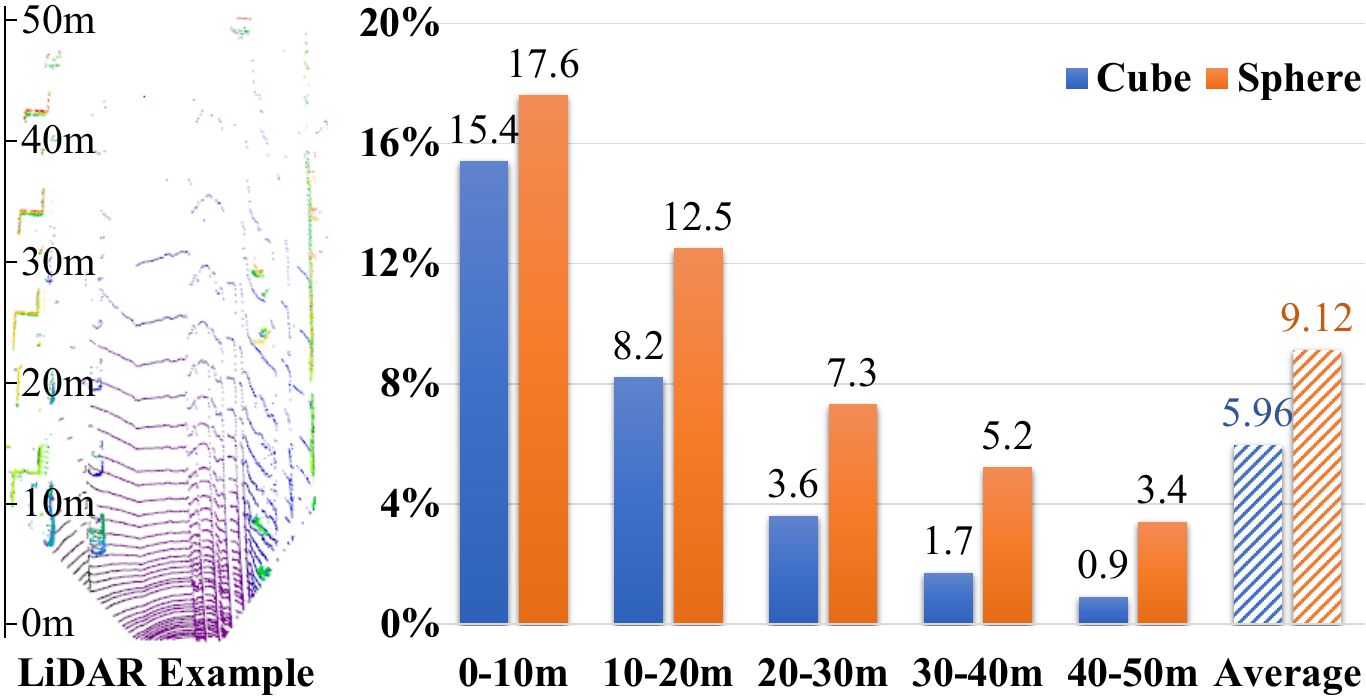}\\
 \caption{Percentage of non-empty units across different distances between cubic and our spherical transformations.}\label{non-empty_cells}
\end{figure}

From the blue bars of Fig.\ref{non-empty_cells} we observe that, \textit{the point clouds exhibit extreme sparsity that is less than 6\%, with their point distributions varying across different distances.} \textbf{To weaken the negative impact of the diverse point distributions, a 3D-to-3D strategy is adopted}. 

(2) \textit{3D-to-3D}: The 3D cubic $\tilde{\mathbf{F}}_{xyz}^i$ is re-projected into the 3D spherical coordinate by $\mathcal{P}_{sph}$ that produces $\mathbf{F}_{r\theta\varphi}^i$. Then, a distance-aware spherical convolution (DASC) is applied to create the 3D spherical feature $\tilde{\mathbf{F}}_{r\theta\varphi}^i$, which refines the geometry in the more compact space: 
\begin{align}
    &\mathbf{F}_{r\theta\varphi}^i=\mathcal{P}_{sph}(\tilde{\mathbf{F}}_{xyz}^i),\label{eq_step2-1}\\
    &\tilde{\mathbf{F}}_{r\theta\varphi}^i=h_{dasc}(\mathbf{F}_{r\theta\varphi}^i),\label{eq_step2-2}
\end{align}
where $h_{dasc}\left( \cdot \right)$ refers to the DASC function (see Eq.~\ref{eq_dasc}). 

From the orange bars of Fig.~\ref{non-empty_cells} we discover that, our 3D-to-3D strategy can better balance the varying point distributions, especially over long distances. After extracting the rich geometric structures in 3D space, \textbf{we employ a 3D-to-2D tactic to further densify the sparse depth}. 

(3) \textit{3D-to-2D}: The 3D feature $\tilde{\mathbf{F}}_{r\theta\varphi}^i$ is projected into 2D space to update the initial 2D $\mathbf{F}_t^i$, $\mathbf{F}_s^i$, and $\mathbf{F}_f^i$ with 2D convolutions $h_{2c}$, yielding new 2D TPV features:
\begin{equation}\label{eq_step3}
\tilde{\mathbf{F}}_t^i,\ \tilde{\mathbf{F}}_s^i,\ \tilde{\mathbf{F}}_f^i=h_{2c}(\mathcal{P}_{tpv}(\mathcal{P}_{sph}^{-1}(\tilde{\mathbf{F}}_{r\theta\varphi}^i))).
\end{equation}

\textit{In the TPV Fusion process, the 2D decoder layers generate an increased number of valid pixels, which enriches the 3D process with a higher density of points. Concurrently, the 3D process captures geometry and feeds it back into the 2D process. These two processes are complementary. }

Particularly, at the output ends of the three TPV subnetworks, we employ three 2D convolutions to predict coarse TPV depth results, obtaining: 
\begin{equation}\label{eq_coarse_predictions}
\mathbf{O}_t,\ \mathbf{O}_s,\ \mathbf{O}_f=h_{2c}(\tilde{\mathbf{F}}_t^4), \ h_{2c}(\tilde{\mathbf{F}}_s^4), \ h_{2c}(\tilde{\mathbf{F}}_f^4).
\end{equation}

\noindent\textbf{Distance-Aware Spherical Convolution.} Given the 3D input $\mathbf{F}_{r\theta\varphi}^i$ in Eq.~\ref{eq_step2-1}, it is sliced by $\mathcal{S}$ into different spherical subareas $\mathbf{A}_{sph}=\{\mathbf{A}_{sph}^1, \cdots ,\mathbf{A}_{sph}^j\}$, each with larger volume $|\mathbf{A}_{sph}^j|$ as the distance $d$ increases, \textit{i.e.}, $|\mathbf{A}^j|\propto d$. Then, these spherical subareas are flattened by $\mathcal{F}$\footnote{Equirectangular projection (ERP) in DUL \cite{yan2023distortion}} into cubic shapes $\mathbf{A}_{cub}=\{\mathbf{A}_{cub}^1, \cdots ,\mathbf{A}_{cub}^j\}$ and filtered by $h_{3c}$, a 3D convolution with kernel $3\times3\times3$ and stride 1. Thus, Eq.~\ref{eq_step2-2} can be written as: 
\begin{equation}\label{eq_dasc}
\tilde{\mathbf{F}}_{r\theta\varphi}^i=\mathcal{F}^{-1}(h_{3c}(\mathcal{F}(\mathcal{S}(\mathbf{F}_{r\theta\varphi}^i))).
\end{equation}

\begin{figure}[t]
 \centering
 \includegraphics[width=0.99\columnwidth]{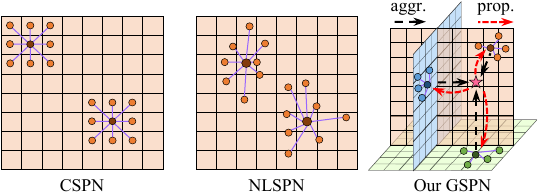}\\
 \caption{Comparison of SPNs~\cite{2018Learning,park2020nonlocal} with different neighbor sets. `aggr.' refers to aggregation while `prop.' indicates propagation.}\label{spn_comparison}
\end{figure}

\subsection{Geometry-Aware Refinement}
\noindent\textbf{Geometric Spatial Propagation Network.} SPNs \cite{liu2017SPN,2018Learning} are widely used to recursively refine the coarse depth $\mathbf{O}_f$. Let $\mathbf{O}_{f(a,b)}$ denotes one pixel at $(a,b)$, while $\mathbf{N}_{f(a,b)}$ indicates its neighbors, one of which is located at $(m,n)$. The propagation of $\mathbf{O}_{f(a,b)}$ at step $(l+1)$ is defined as: 
\begin{equation}\label{eq_spn}
\mathbf{O}_{f(a,b)}^{l+1}=(1-\sum\limits_{m,n}{\omega _{f(a,b)}^{m,n}})\mathbf{O}_{f(a,b)}^{l}+\sum\limits_{m,n}{\omega _{f(a,b)}^{m,n}\mathbf{O}_{f(m,n)}^{l}},
\end{equation}
where $\omega _{f(a,b)}^{m,n}$ is the affinity of pixels at $(a,b)$ and $(m,n)$. 

In Fig.~\ref{spn_comparison}, the key of SPNs is how to search for the neighbor set $\mathbf{N}_{f(a,b)}$. In 2D space, CSPN~\cite{2018Learning} constructs $\mathbf{N}_{f(a,b)}^{CS}$ within a fixed square area excluding the centre pixel, while NLSPN~\cite{park2020nonlocal} deforms it in ${\mathbb{R}}^{H\times W}$ to build $\mathbf{N}_{f(a,b)}^{NL}$: 
\begin{align}
    &\mathbf{N}_{f(a,b)}^{CS}=\{\mathbf{O}_{f(a+u,b+v)} \ | \ u,v\in \{-1,0,1 \}\},\label{eq_cspn}\\
    &\mathbf{N}_{f(a,b)}^{NL}=\{\mathbf{O}_{f(a+u,b+v)} \ | \ u,v\in h_{off}(\mathbf{I},\mathbf{S},a,b) \},\label{eq_nlspn}
\end{align}
where $h_{off}$ learns the offset based on the RGB-D input. 

Differently, given $\mathbf{O}_t^l$, $\mathbf{O}_s^l$, and $\mathbf{O}_f^l$, our GSPN uses the deformable technique $h_{nl}(\cdot)$ in Eqs.~\ref{eq_spn} and \ref{eq_nlspn} to produce the front-view $\mathbf{O}_{f}^{l+1}$, as well as the top-view $\mathbf{O}_{t}^{l+1}$ and side-view $\mathbf{O}_{s}^{l+1}$ in TPV spaces. Then the three views are aggregated in 3D space \cite{huang2023tri,zuo2023pointocc} via projection and MLP. At last, the 3D feature is propagated back to the TPV spaces for refinement: 
\begin{equation}\label{eq_gspn}
\tilde{\mathbf{O}}_t^{l+1},\ \tilde{\mathbf{O}}_s^{l+1},\ \tilde{\mathbf{O}}_f^{l+1}=h_{gspn}(\mathbf{O}_t^{l+1},\mathbf{O}_s^{l+1},\mathbf{O}_f^{l+1}),
\end{equation}
where $h_{gspn}(\cdot)$ refers to $\mathcal{P}_{tpv}(h_{mlp}(\mathcal{P}_{tpv}^{-1}(h_{nl}(\cdot)))$. 


\begin{table*}[t]
\centering
\renewcommand\arraystretch{1.02}
\resizebox{0.995\textwidth}{!}{
\begin{tabular}{l|cc|c|cccc|c}
\toprule
{Method}                           & 2D        & 3D  & Params. (M) $\downarrow$ & RMSE (mm) $\downarrow$     & MAE (mm) $\downarrow$ & iRMSE (1/km) $\downarrow$  & iMAE (1/km) $\downarrow$  & Publication \\ 
\midrule
CSPN \cite{2018Learning}           & \checkmark  &   & 17.4   & 1019.64   & 279.46   & 2.93           & 1.15           & ECCV 2018 \\
S2D \cite{ma2018self}              & \checkmark  &   & 26.1   & 814.73    & 249.95   & 2.80           & 1.21           & ICRA 2019 \\
NConv \cite{2020Confidence}        & \checkmark  &   & \textbf{0.36}   & 829.98    & 233.26   & 2.60           & 1.03           & PAMI 2020 \\
CSPN++ \cite{Cheng2020CSPN++}        & \checkmark  &   & 26.0   & 743.69    & 209.28   & 2.07           & 0.90           & AAAI 2020 \\
NLSPN \cite{park2020nonlocal}      & \checkmark  &   & 25.8   & 741.68    & 199.59   & 1.99           & 0.84           & ECCV 2020 \\
GuideNet \cite{tang2020learning}   & \checkmark  &   & 73.5   & 736.24    & 218.83   & 2.25           & 0.99           & TIP 2020 \\
TWISE \cite{imran2021depth}        & \checkmark  &   & \textcolor{blue}{1.45}   & 840.20    & 195.58   & 2.08           & \textcolor{blue}{0.82}           & CVPR 2021 \\
FCFRNet \cite{liu2021fcfr}         & \checkmark  &   & 50.6   & 735.81    & 217.15   & 2.20           & 0.98           & AAAI 2021 \\
PENet \cite{hu2020PENet}           & \checkmark  &   & 131.5  & 730.08    & 210.55   & 2.17           & 0.94           & ICRA 2021 \\
DySPN \cite{lin2022dynamic}        & \checkmark  &   & 26.3   & 709.12    & 192.71   & 1.88           & \textcolor{blue}{0.82}           & AAAI 2022 \\
RigNet \cite{yan2022rignet}        & \checkmark  &   & 65.2   & 712.66    & 203.25   & 2.08           & 0.90           & ECCV 2022 \\ 
CFormer \cite{zhang2023cf}         & \checkmark  &   & 83.5   & 708.87    & 203.45   & 2.01           & 0.88           & CVPR 2023 \\
RigNet++ \cite{yan2023rignet++}    & \checkmark  &   & 19.9   & 710.85    & 202.45   & 2.01           & 0.89           & arXiv 2023 \\
LRRU \cite{wang2023lrru}           & \checkmark  &   & 21.0   & \textcolor{blue}{696.51}    & 189.96   & 1.87           & \textbf{0.81}  & ICCV 2023 \\ 
\midrule
DepthNormal \cite{Xu2019Depth} & \checkmark & \checkmark & $\sim$ 40 & 777.05    & 235.17   & 2.42           & 1.13           & ICCV 2019 \\
FuseNet$^\star$ \cite{chen2019learning}    & \checkmark & \checkmark & 1.9 & 752.88    & 221.19   & 2.34           & 1.14           & ICCV 2019 \\
DLiDAR$^\star$ \cite{Qiu_2019_CVPR}        & \checkmark & \checkmark& 53.4   & 758.38    & 226.50   & 2.56           & 1.15           & CVPR 2019 \\
ACMNet \cite{zhao2021adaptive}     & \checkmark & \checkmark & 4.9 & 744.91    & 206.09   & 2.08           & 0.90           & TIP 2021 \\
PointFusion\cite{huynh2021boosting}& \checkmark & \checkmark & 8.7 & 741.90    & 201.10   & 1.97           & 0.85           & ICCV 2021 \\
GraphCSPN \cite{liu2022graphcspn}  & \checkmark & \checkmark & 26.4 & 738.41    & 199.31   & 1.96           & 0.84           & ECCV 2022 \\
BEV@DC \cite{zhou2023bev}           & \checkmark & \checkmark & 30.8 & 697.44    & \textcolor{blue}{189.44}   & \textcolor{blue}{1.83}           & \textcolor{blue}{0.82}           & CVPR 2023 \\
PointDC \cite{yu2023aggregating}   & \checkmark & \checkmark & 25.1 & 736.07    & 201.87   & 1.97           & 0.87           & ICCV 2023 \\
\midrule
\textbf{TPVD (ours)}               & \checkmark & \checkmark & 31.2 & \textbf{693.97} & \textbf{188.60} & \textbf{1.82} & \textbf{0.81} & CVPR 2024 \\ 
\bottomrule
\end{tabular}
}
\vspace{-7pt}
\caption{Quantitative results on KITTI online depth completion leaderboard. 2D and 3D refer to models that involve 2D and 3D representations, respectively. $^\star$ denotes models that involve additional training data. The \textbf{best} and the \textcolor{blue}{second best} metrics are highlighted.}
\label{tab_kitti}
\vspace{-8pt}
\end{table*}

\section{TOFDC}
\begin{figure}[t]
  \centering
  \includegraphics[width=0.98\columnwidth]{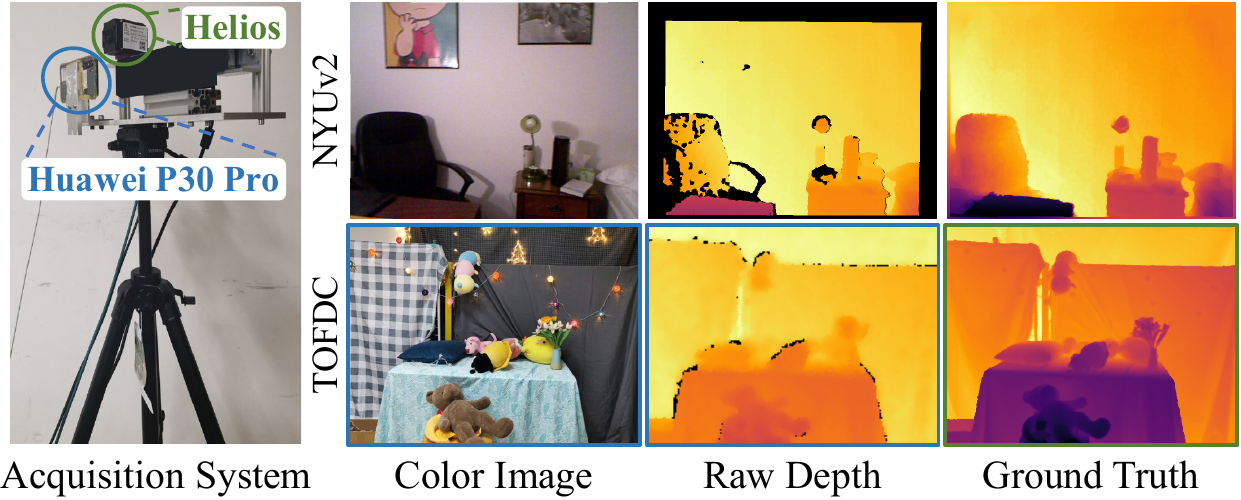}\\
  \caption{Acquisition system (left) and data comparison (right).}\label{collection_raw_data}
\end{figure}

\begin{figure}[t]
  \centering
  \includegraphics[width=0.8\columnwidth]{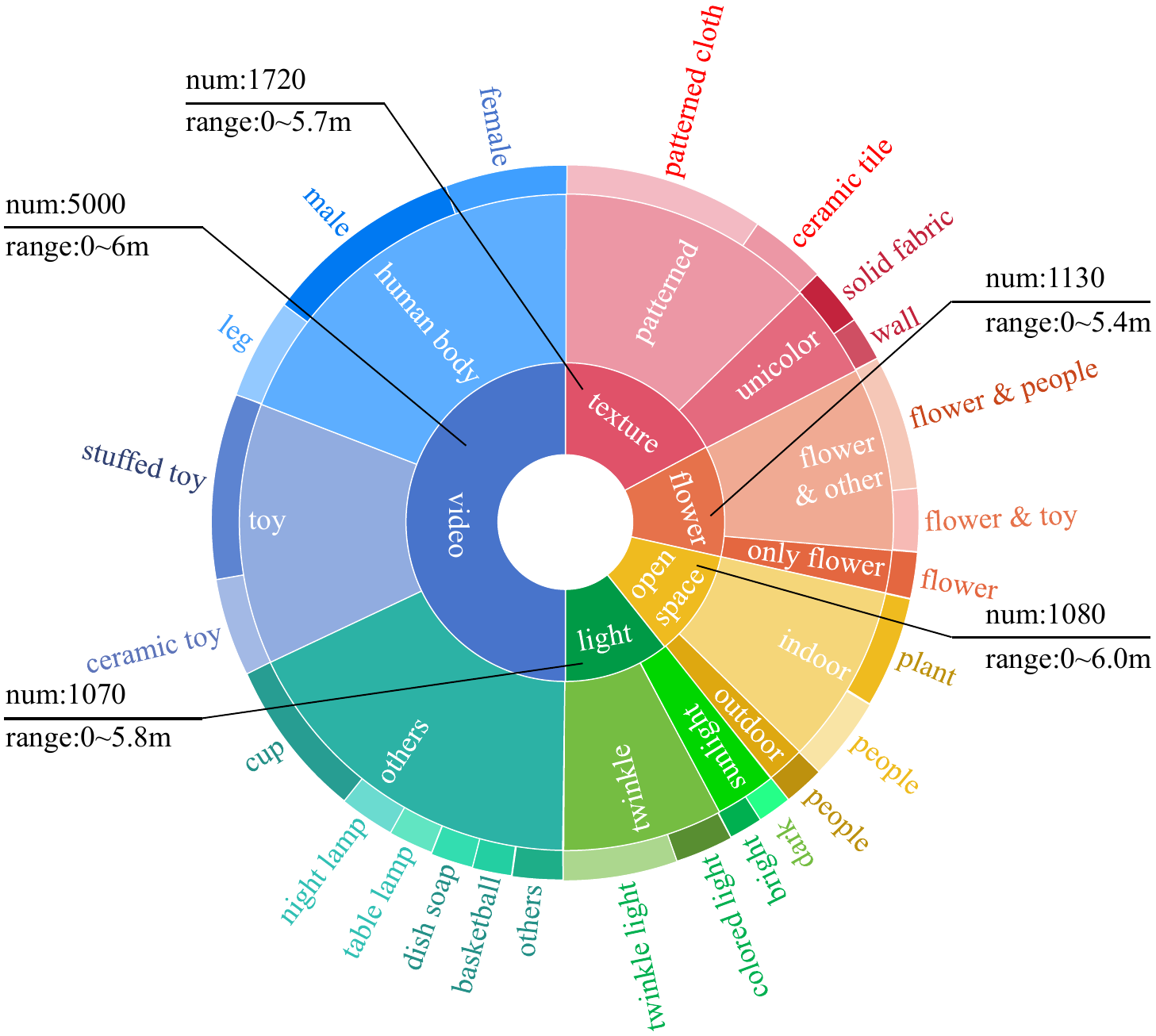}\\
  \vspace{-8pt}
  \caption{Distribution of different scenarios in our TOFDC.
  }\label{tofdc_statistic_scene}
  \vspace{-6pt}
\end{figure}

\cref{collection_raw_data} shows the data acquisition system and data comparison between NYUv2 \cite{silberman2012indoor} and our TOFDC. The system consists of the Huawei P30 Pro (for color image and raw depth) and Helios (for ground truth depth). We find that the depth of TOFDC is much denser than NYUv2. 
\cref{tofdc_statistic_scene} shows the distribution of different scenarios in TOFDC, which stands for texture, flower, light, open space, and video, and we have collected \textbf{10,000} RGB-D pairs from these scenarios in total. Please see our appendix for more details.

\section{Experiments} 
\subsection{Datasets}
\noindent\textbf{TOFDC} is collected by the TOF sensor and RGB camera of a Huawei P30 Pro, which covers various scenes such as texture, flower, body, and toy, under different lighting conditions and in open space. It has \textbf{10k} $512\times 384$ RGB-D pairs for training and \textbf{560} for evaluation. The ground truth depth maps are captured by the Helios TOF camera. 

\noindent\textbf{KITTI} dataset \cite{Uhrig2017THREEDV} contains 86k training samples, 1k selected validation samples, and 1k online test samples without ground truths. The depth data is captured by a 64-line LiDAR sensor. Following \cite{tang2020learning,yan2023desnet,lin2023dyspn}, the RGB-D pairs are bottom center cropped from $1216\times 352$ to $1216\times 256$, as there are no valid LiDAR values near top 100 pixels. 

\noindent\textbf{NYUv2} dataset \cite{silberman2012indoor} consists of paired RGB-D from 464 indoor scenes, where the depth maps are acquired by Microsoft Kinect. We train our model with 50K samples and test it on the official 654 samples. Following \cite{yan2022rignet,liu2023mff,zhou2023bev,yu2023aggregating}, we first downsample the RGB-D pairs from $640\times 480$ to $320\times 240$, and then center crop it to $304\times 228$.

\noindent\textbf{SUN RGBD} dataset \cite{song2015sun} is selected from several indoor RGB-D datasets \cite{silberman2012indoor,janoch2013category,xiao2013sun3d}. We use 555 samples captured by Kinect V1 and 3,389 samples captured by Asus Xtion camera for cross-dataset evaluation, where we employ the same pre-processing step as that on the NYU2 dataset.

 \begin{figure*}[t]
  \centering
  \includegraphics[width=1.95\columnwidth]{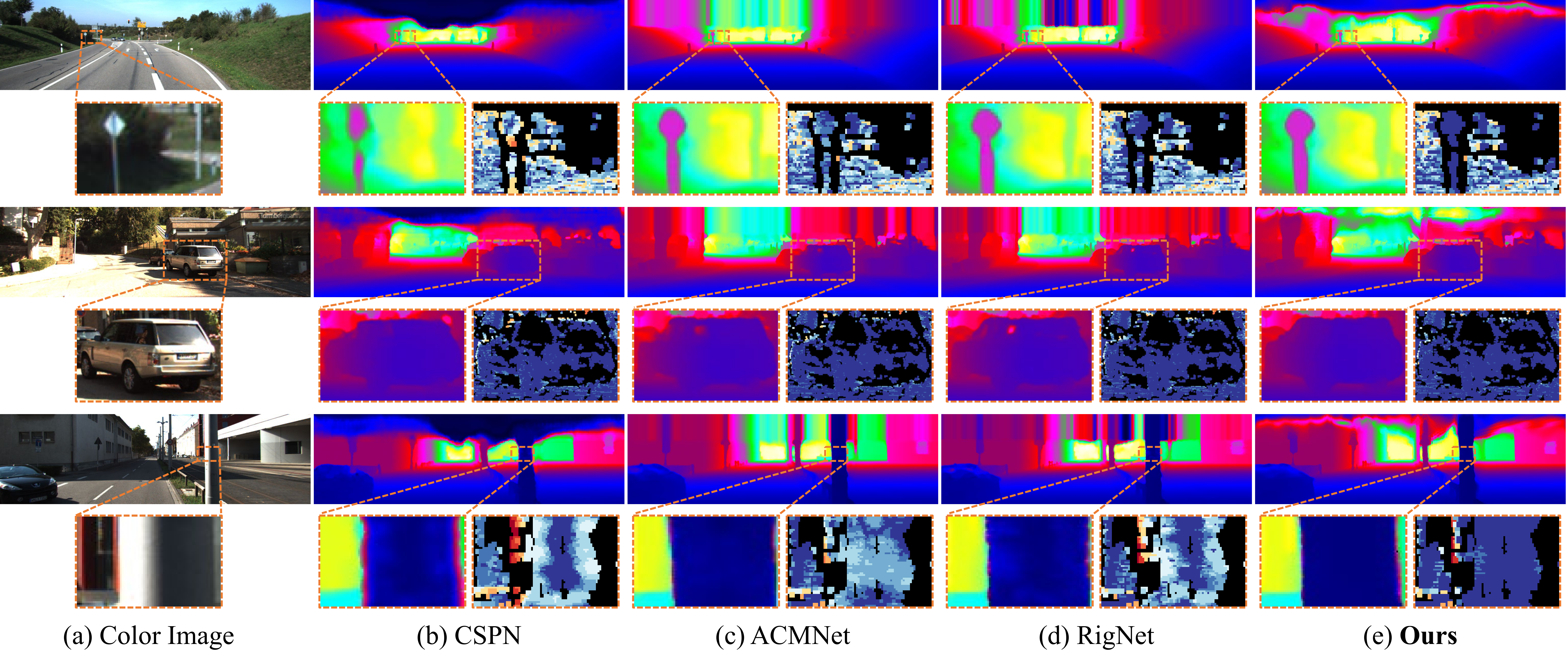}\\
  \caption{Qualitative results on KITTI depth completion benchmark, including (b) CSPN \cite{2018Learning}, (c) ACMNet \cite{zhao2021adaptive}, (d) RigNet \cite{yan2022rignet}, and (e) our TPVD method. The zoomed-in regions and their corresponding error maps (the darker, the better) show more fine-grained differences.}\label{kitti_vis}
\end{figure*}

\begin{table}[t]
\centering
\renewcommand\arraystretch{1.02}
\resizebox{0.475\textwidth}{!}{
\begin{tabular}{l|c|c|ccc}
\toprule
Method  & Params. $\downarrow$  & FLOPs $\downarrow$  & Train $\uparrow$  & Test $\uparrow$ \\ 
\midrule 
ACMNet \cite{zhao2021adaptive}  & \textbf{4.9} M  & 544 G   & 2.72 FPS   & 4.20 FPS  \\
BEV@DC \cite{zhou2023bev}  & \textcolor{blue}{26.9} M  & \textcolor{blue}{462} G  & \textcolor{blue}{3.01} FPS   & \textcolor{blue}{7.87} FPS  \\
\textbf{TPVD (ours)}  & 31.2 M  & \textbf{328} G  & \textbf{3.63} FPS  & \textbf{8.82} FPS \\
\bottomrule
\end{tabular}
}
\caption{Train \& test speed comparison on KITTI validation set.}\label{tab_speed}
\end{table}

\subsection{Comparison with State-of-the-arts} 

\textbf{Outdoor KITTI.} We first evaluate the proposed TPVD on KITTI depth completion benchmark that is ranked by RMSE. The top part of Tab.~\ref{tab_kitti} lists the results of 2D based methods while the bottom part reports those of 2D-3D joint approaches. On the whole, TPVD ranks \textbf{1st} among all the methods in four evaluation metrics at the time of submission, including RMSE, MAE, iRMSE, and iMAE. For example, TPVD is 15.98 mm superior to the five latest researches on average, \textit{i.e.}, CFormer~\cite{zhang2023cf}, BEV@DC~\cite{zhou2023bev}, LRRU~\cite{wang2023lrru}, PointDC~\cite{yu2023aggregating}, and RigNet++~\cite{yan2023rignet++}. Among the 2D-3D joint counterparts, compared with the lightweight FuseNet~\cite{chen2019learning}, ACMNet~\cite{zhao2021adaptive}, and PointFusion~\cite{huynh2021boosting}, the errors of TPVD are significantly lower, \textit{e.g.}, averagely by 52.59 mm in RMSE and 20.86 mm in MAE. In contrast to those 2D-3D joint methods with similar or larger parameters, TPVD still performs better. 
Fig.~\ref{kitti_vis} shows the visual comparison with CSPN~\cite{2018Learning}, ACMNet~\cite{zhao2021adaptive}, and RigNet~\cite{yan2022rignet}. While they produce visually good predictions in general, TPVD can recover more accurate shapes and boundaries. The zoom-in error maps further indicate the superiority.

In addition, Tab.~\ref{tab_speed} lists the complexity and speed comparison of the 2D-3D joint ACMNet~\cite{zhao2021adaptive}, BEV@DC~\cite{zhou2023bev}, and TPVD. We observe that, despite ACMNet having fewer parameters, its graph model is more complex and requires about twice as many FLOPs as ours. Consequently, ACMNet suffers from slower training and testing speeds. Differently, the LiDAR stream of BEV@DC is removed in testing phase, improving the testing speed from 3.01 FPS to 7.87 FPS. Different from them, our TPV design is computation-friendly though the parameters are slightly higher. The FLOPs is 134 G lower than the second-best BEV@DC, contributing to faster training and testing speeds.

\begin{table}[t]
\centering
\renewcommand\arraystretch{1.02}
\resizebox{0.478\textwidth}{!}{
\begin{tabular}{l|ccccc}
\toprule
Method          & RMSE (m) $\downarrow$      & REL $\downarrow$     &${\delta }_{1}$  $\uparrow$ & ${\delta }_{{2}}$ $\uparrow$ & ${\delta }_{{3}}$  $\uparrow$ \\ 
\midrule 
CSPN \cite{2018Learning}         & 0.117    & 0.016   & 99.2  & 99.9  & 100.0 \\
FCFRNet \cite{liu2021fcfr}       & 0.106    & 0.015   & 99.5  & 99.9  & 100.0 \\
GuideNet \cite{tang2020learning} & 0.101    & 0.015   & 99.5  & 99.9  & 100.0 \\
NLSPN \cite{park2020nonlocal}    & 0.092    & 0.012   & 99.6  & 99.9  & 100.0 \\ 
DySPN \cite{lin2022dynamic}      & 0.090    & 0.012   & 99.6  & 99.9  & 100.0 \\
CFormer \cite{zhang2023cf}       & 0.091    & 0.012   & 99.6  & 99.9  & 100.0 \\
RigNet \cite{yan2022rignet}      & 0.090    & 0.013   & 99.6  & 99.9  & 100.0  \\
LRRU \cite{wang2023lrru}         & 0.091    & \textcolor{blue}{0.011}   & 99.6  & 99.9  & 100.0  \\
\midrule
DLiDAR \cite{Qiu_2019_CVPR}      & 0.115    & 0.022   & 99.3  & 99.9  & 100.0 \\
ACMNet \cite{zhao2021adaptive}   & 0.105    & 0.015   & 99.4  & 99.9  & 100.0 \\
GraphCSPN \cite{liu2022graphcspn}& 0.090    & 0.012   & 99.6  & 99.9  & 100.0  \\ 
BEV@DC \cite{zhou2023bev}         & \textcolor{blue}{0.089}    & 0.012   & 99.6  & 99.9  & 100.0 \\ 
PointDC \cite{yu2023aggregating} & \textcolor{blue}{0.089}    & 0.012   & 99.6  & 99.9  & 100.0  \\
\textbf{TPVD (ours)}             & \textbf{0.086}    & \textbf{0.010}   & \textbf{99.7}  & \textbf{99.9}  & \textbf{100.0}  \\ 
\bottomrule
\end{tabular}
}
\caption{Quantitative comparison on NYUv2 dataset. The second row shows the results of 2D based methods, whilst the third row illustrates those of 2D-3D joint approaches.}\label{tab_nyuv2}
\end{table}

 \begin{figure}[t]
  \centering
  \includegraphics[width=0.993\columnwidth]{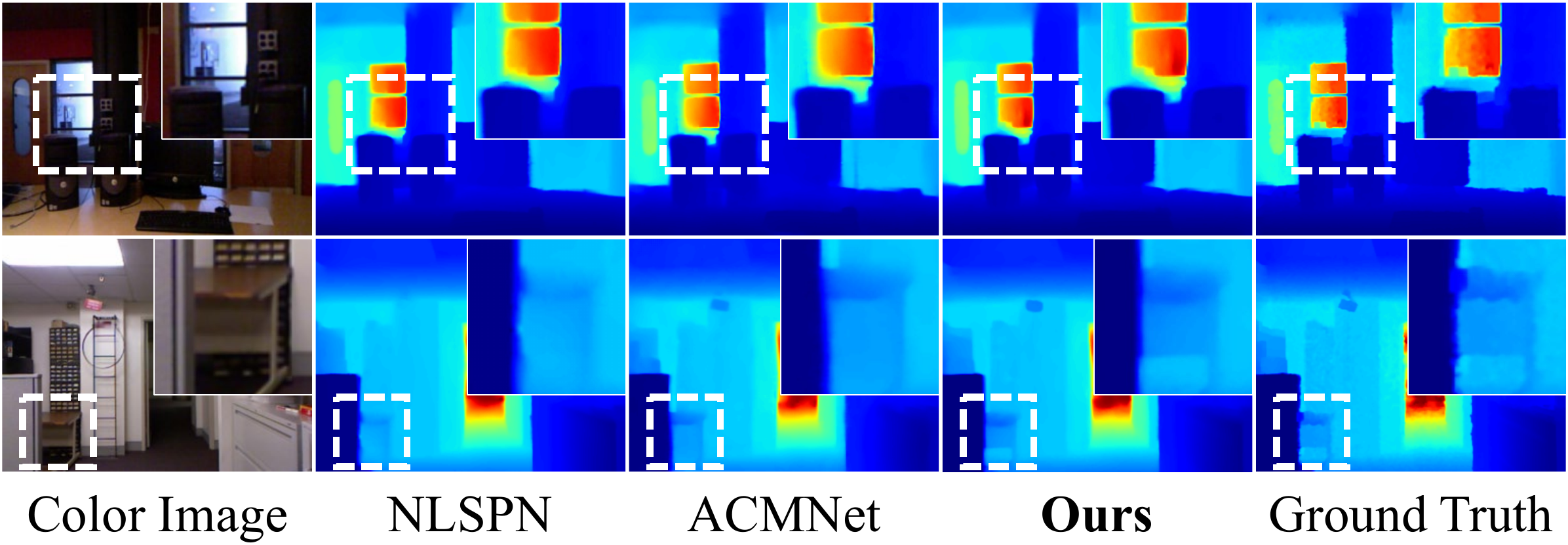}\\
  \caption{Visual comparison of NLSPN \cite{park2020nonlocal}, ACMNet \cite{zhao2021adaptive}, and our TPVD method on NYUv2 dataset.}\label{nyu_vis}
\end{figure}

\noindent \textbf{Indoor NYUv2.} To verify the effectiveness of TPVD on indoor scenes, following~\cite{tang2020learning,park2020nonlocal,yan2022rignet}, we train TPVD on NYUv2 dataset with 500 sampling depth pixels. As listed in Tab.~\ref{tab_nyuv2}, the top and bottom parts refer to 2D based and 2D-3D joint categories, respectively. We can observe that TPVD still achieve the best performance in all five metrics. Particularly, compared to previous state-of-the-art methods~\cite{zhang2023cf,zhou2023bev,wang2023lrru,yu2023aggregating} that are only 1 mm superior in RMSE to concurrent works, our TPVD attains 3 mm improvement again. Meanwhile, the REL is reduced by 20\% over the latest 2D-3D joint BEV@DC~\cite{zhou2023bev} and PointDC~\cite{yu2023aggregating}. Fig.~\ref{nyu_vis} shows that TPVD succeeds in restoring detailed structures.

 \begin{figure}[t]
  \centering
  \includegraphics[width=0.995\columnwidth]{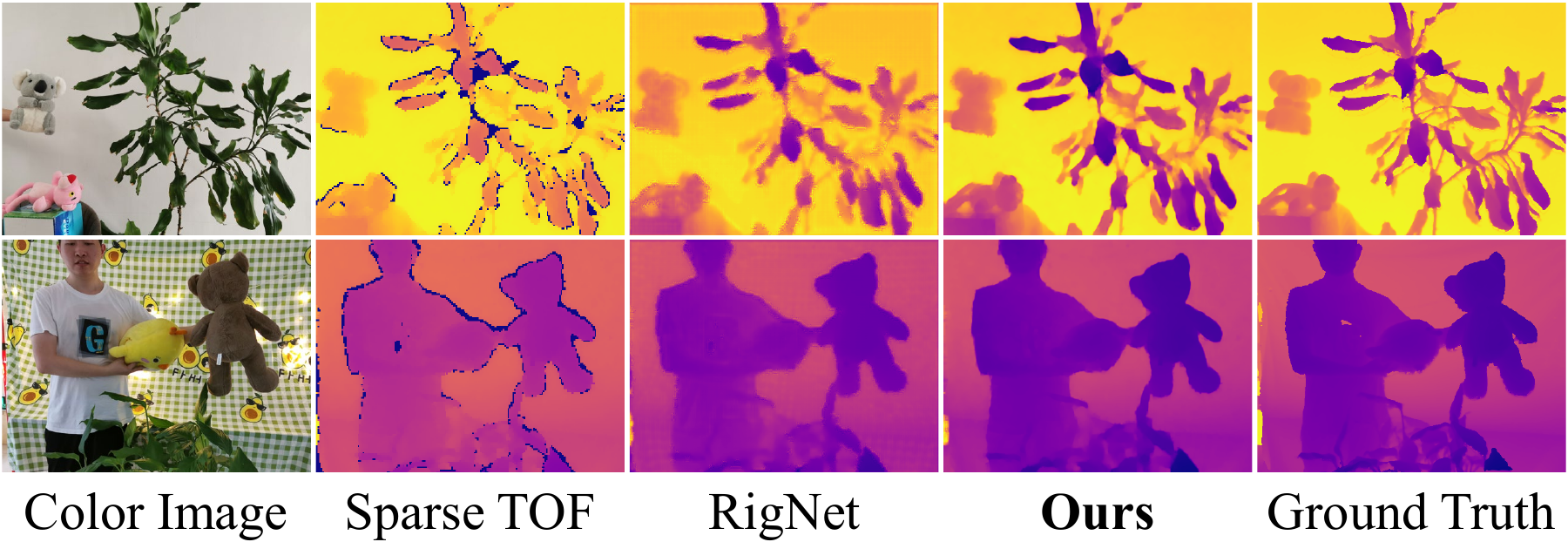}\\
  \caption{Visual results of RigNet \cite{yan2022rignet} and TPVD on TOFDC.}\label{tofdc_vis}
\end{figure}

\begin{table}[t]
\centering
\renewcommand\arraystretch{1.02}
\resizebox{0.478\textwidth}{!}{
\begin{tabular}{l|ccccc}
\toprule
Method          & RMSE (m) $\downarrow$      & REL $\downarrow$     &${\delta }_{1}$  $\uparrow$ & ${\delta }_{{2}}$ $\uparrow$ & ${\delta }_{{3}}$  $\uparrow$ \\ 
\midrule 
CSPN \cite{2018Learning}          & 0.224  & 0.042  & 94.5  & 95.3  & 96.5  \\
FusionNet \cite{vangansbeke2019}  & 0.116  & 0.024  & 98.3  & 99.4  & 99.7  \\
GuideNet \cite{tang2020learning}  & 0.146  & 0.030  & 97.6  & 98.9  & 99.5  \\
ENet \cite{hu2020PENet}           & 0.231  & 0.061  & 94.3  & 95.2  & 97.4  \\
PENet \cite{hu2020PENet}          & 0.241  & 0.043  & 94.6  & 95.3  & 95.5  \\
NLSPN \cite{park2020nonlocal}     & 0.174  & 0.029  & 96.4  & 97.9  & 98.9  \\
CFormer \cite{zhang2023cf}        & 0.113  & 0.029 & 99.1   & 99.6  & 99.9  \\
RigNet \cite{yan2022rignet}       & 0.133  & 0.025  & 97.6  & 99.1  & 99.7  \\
\midrule
GraphCSPN \cite{liu2022graphcspn} & 0.253  & 0.052  & 92.0  & 96.9  & 98.7  \\ 
PointDC \cite{yu2023aggregating}  & \textcolor{blue}{0.109}  & \textcolor{blue}{0.021}  & 98.5  & 99.2  & 99.6  \\
\textbf{TPVD (ours)}              & \textbf{0.092}    & \textbf{0.014}   & \textbf{99.1}  & \textbf{99.6}  & \textbf{99.9}  \\ 
\bottomrule
\end{tabular}
}
\caption{Quantitative comparison on our new TOFDC dataset.}\label{tab_tofdc}
\end{table}

\noindent \textbf{Indoor TOFDC.} To further test our TPVD, we implement it on the new TOFDC dataset that is collected by consumptive TOF sensors. As reported in Tab.~\ref{tab_tofdc}, 2D based and 2D-3D joint methods are divided into the top part and the bottom part, severally. We discover that TPVD outperforms the 2D-3D joint approaches by a large margin. For example, it reduces the RMSE by 15.6\% and REL by 33.3\% against the second best PointDC~\cite{yu2023aggregating}. Also, compared with the best 2D based CFormer~\cite{zhang2023cf}, TPVD is 21 mm superior in RMSE, which is a considerable improvement for indoor scenes. Fig.~\ref{tofdc_vis} reveals that TPVD can predict high-quality dense depth results with clearer and sharper structures.

\subsection{Generalization Capability}
\noindent \textbf{Depth-Only Input.} For depth completion task, the auxiliary color images may not always be accessible or dependable, for instance, when the camera malfunctions or when lighting conditions are extremely poor, such as at night. Consequently, we assess our TPVD under a depth-only setting, and compare it with previous methods in Tab.~\ref{tab_depth_only}. Compared to the depth-only IP\_Basic~\cite{ku2018defense}, S2D~\cite{ma2018self}, FusionNet~\cite{vangansbeke2019}, and LRRU~\cite{wang2023lrru}, TPVD achieves the lowest RMSE and MAE, surpassing the second best by 8.8 mm and 4.3 mm, respectively. Furthermore, the MAE of TPVD is significantly superior to that of IR by 65.8 mm though the RMSE is higher. It’s noteworthy that TPVD solely takes sparse depth as input, whereas IR uses color images as supervisory signals during training. These analyses indicate that TPVD can work well without image guidance.

\begin{table}[t]
\centering
\renewcommand\arraystretch{1.02}
\resizebox{0.478\textwidth}{!}{
\begin{tabular}{l|c|cc}
\toprule
Method    & Specialty  & RMSE (mm) $\downarrow$            & MAE (mm) $\downarrow$ \\ 
\midrule
IP\_Basic~\cite{ku2018defense}    & params. free & 1350.9  & 305.4  \\
S2D~\cite{ma2018self}             & depth only   & 985.1   & 286.5  \\
FusionNet~\cite{vangansbeke2019}  & depth only   & 995.0   & 268.0  \\
IR~\cite{2020FromLu}              & RGB assisted & \textbf{914.7} & 297.4 \\
LRRU~\cite{wang2023lrru}          & depth only   & 957.4   & \textcolor{blue}{235.9}  \\
\textbf{TPVD (ours)}              &  depth only  & \textcolor{blue}{948.6}  & \textbf{231.6}  \\
\bottomrule
\end{tabular}
}
\caption{Depth-only comparison on KITTI validation split.}\label{tab_depth_only}
\end{table}

\begin{figure}[t]
 \centering
 \includegraphics[width=0.994\columnwidth]{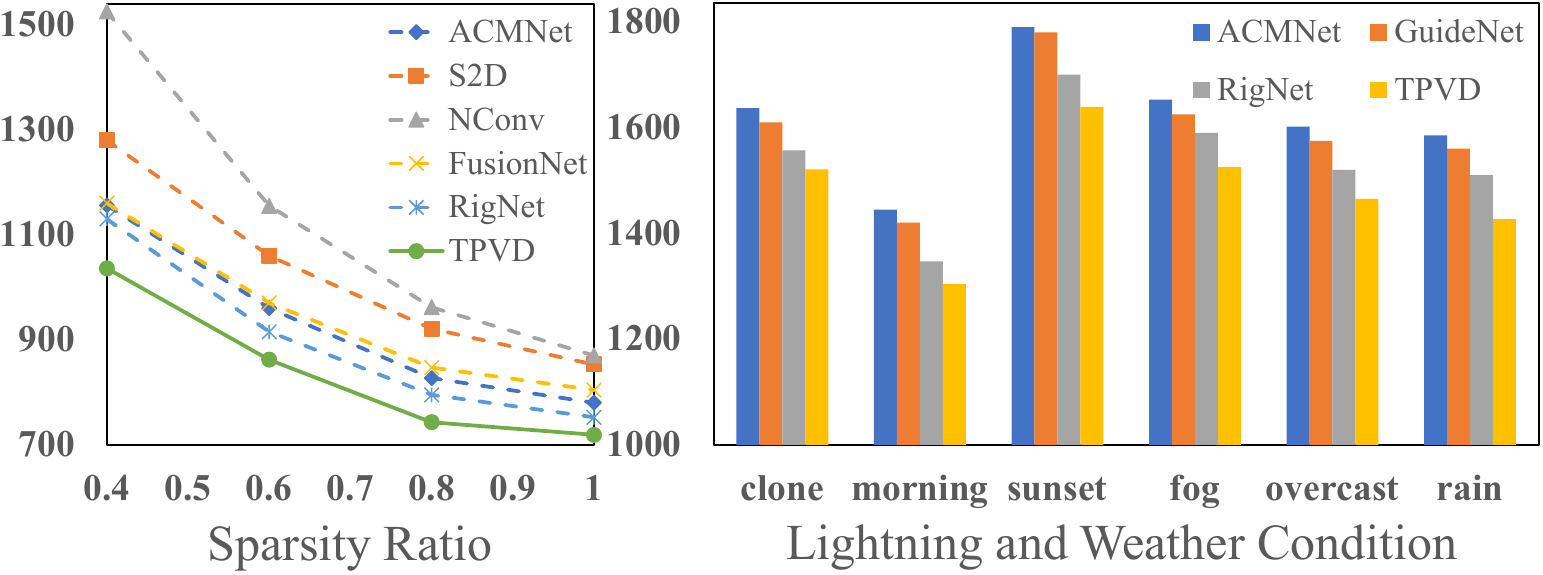}\\
 \caption{RMSE (mm) comparison under different sparsity ratios on KITTI validation split (left), and diverse lightning and weather conditions on VKITTI dataset \cite{gaidon2016virtual} (right).}\label{sparsity_weather}
\end{figure}

\noindent \textbf{Number of Valid Points.} We compare the proposed TPVD with five well-known methods with available codes, \textit{i.e.}, S2D~\cite{ma2018self}, NConv~\cite{2020Confidence}, FusionNet~\cite{vangansbeke2019}, ACMNet~\cite{zhao2021adaptive}, and RigNet~\cite{yan2022rignet}. Following \cite{ma2018self,yan2022rignet}, we first conduct uniform sampling to produce sparser depth input with ratios (0.4, 0.6, 0.8, 1), where the raw sparsity corresponds to the sampling ratio 1. Then we retrain all the approaches on KITTI and test them on the official validation split. As shown in the left of Fig.~\ref{sparsity_weather}, our TPVD achieves considerable superiority against other methods under all sparsity ratios. These results demonstrate that the proposed TPVD still can perform well even with complex data input.

\noindent \textbf{Lightning and Weather Condition.} KITTI dataset is collected on sunny days~\cite{yan2022rignet}, whose lightning is almost unchanging and the weather is satisfactory. However, in real-world environments, both factors can be quite complex and pose significant challenges for autonomous driving application. Therefore, we first fine-tune our TPVD (pretrained on KITTI) on ``clone" of VKITTI~\cite{gaidon2016virtual} and then test it on the other scenes with various lightning and weather conditions. In Fig.~\ref{sparsity_weather} (right), we compare TPVD with GuideNet~\cite{tang2020learning}, ACMNet~\cite{zhao2021adaptive}, and RigNet~\cite{yan2022rignet}. Obviously, our method surpasses the three approaches consistently on morning, sunset, fog, overcast, and rain scenes. It indicates that TPVD can tackle complex lightning and weather conditions. 

See Supp. for cross-dataset evaluation on SUN RGBD.

\subsection{Ablation Studies}
\noindent \textbf{TPVD Designs.} Tab.~\ref{tab_ablation_on_tpvd} lists the ablation results on KITTI validation split. The baseline model, TPVD-i, solely incorporates the front-view depth. When introducing the top view depth in TPVD-ii, the RMSE decreases from 763.56 mm to 755.15 mm. Building upon TPVD-ii, TPVD-iii integrates the depth of the front, top, and side views, providing comprehensive initial 3D geometry and leading to an improvement of 5.87 mm in RMSE. In TPVD-iv, the application of the proposed DASC further reduces the RMSE by 13.81 mm, marking a significant enhancement. Those improvements in TPVD-ii, iii, and iv over the baseline are primarily attributed to the increased 3D geometric awareness. Lastly, TPVD-v surpasses TPVD-iv by 16.67 mm in RMSE and 3.11 mm in MAE, underscoring the efficacy of GSPN in generating consistent fine-grained geometry through propagation in TPV spaces. In brief, each proposed component contributes positively to the performance of the baseline.

\begin{table}[t]
\centering
\renewcommand\arraystretch{1.02}
\resizebox{0.476\textwidth}{!}{
\begin{tabular}{l|cccc|c|cc}
\toprule
\multirow{2}{*}{TPVD} & \multicolumn{4}{c|}{TPV Fusion} & \multirow{2}{*}{GSPN} & RMSE & MAE \\ \cline{2-5} 
& front  & top  & side  & DASC  &  & (mm) & (mm)    \\ \midrule
i   & \checkmark  &             &             &             &             & 763.56  & 197.82  \\
ii  & \checkmark  & \checkmark  &             &             &             & 755.14  & 194.85  \\
iii & \checkmark  & \checkmark  & \checkmark  &             &             & 749.38  & 192.51  \\
iv  & \checkmark  & \checkmark  & \checkmark  & \checkmark  &             & 
735.57 & 190.26  \\
v   & \checkmark  & \checkmark  & \checkmark  & \checkmark  & \checkmark  & \textbf{718.90}  & \textbf{187.15}  \\ \bottomrule
\end{tabular}
}
\vspace{-5pt}
\caption{Ablation studies of our TPVD on KITTI validation split.}
\label{tab_ablation_on_tpvd}
\vspace{-3pt}
\end{table}

\begin{figure}[t]
 \centering
 \includegraphics[width=0.998\columnwidth]{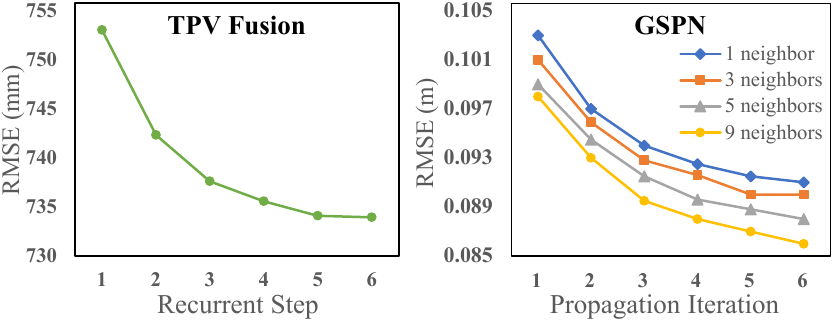}\\
 \caption{Ablation studies of TPV Fusion on KITTI validation split (left), and GSPN on NYUv2 test set (right).}\label{fig_tpvf_gspn}
\end{figure}

\noindent \textbf{TPV Fusion.} The left side of Fig.~\ref{fig_tpvf_gspn} presents the ablation of TPV Fusion with varying recurrent steps on KITTI validation split. Overall, it can be observed that the error decreases as the recurrent steps increases. For instance, the second step improves upon the first step by approximately 9 mm. However, these limited recurrent steps do not provide sufficient geometric aggregation. Moreover, when the number of steps exceeds 4, the improvement becomes negligible. Consequently, we set the recurrent step to 4 to strike a balance between efficiency and effectiveness.

\begin{figure}[t]
 \centering
 \includegraphics[width=0.998\columnwidth]{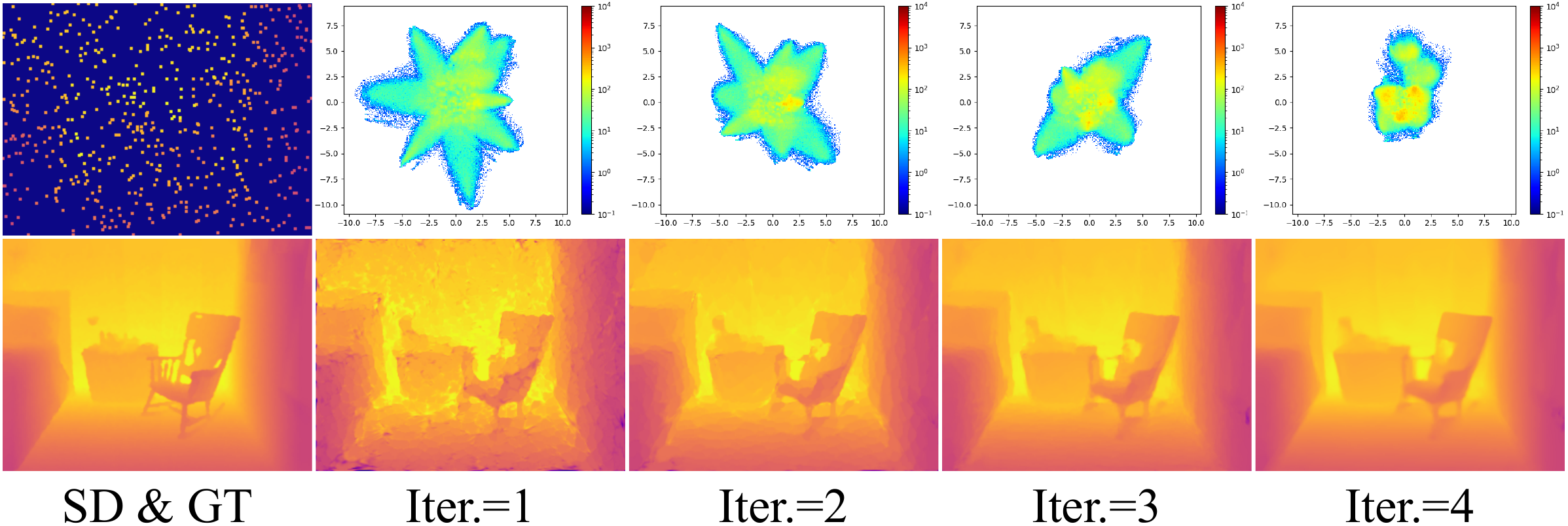}\\
 \caption{Visual process of GSPN on NYUv2. 1st row: receptive fields of kernels in top-view sparse depth. 2nd row: dense results.}\label{fig_gspn_kernel_vis}
\end{figure}

\noindent \textbf{GSPN.} The right side of Fig.~\ref{fig_tpvf_gspn} ablates GSPN on NYUv2. We find that, (1) a larger number of neighbors leads to lower errors, \textit{e.g.}, the RMSE of 9 neighbors is on average 3.3 mm better than that of 5 neighbors. (2) The performance improves as the iteration increases. When the number is 9 and iteration is 6, GSPN achieves the best result. For efficiency-effectiveness trade-off, we set the neighbor and iteration to 9 and 4, respectively. Fig.~\ref{fig_gspn_kernel_vis} shows that with each successive iteration, GSPN progressively produces denser depth with more precise geometry. Furthermore, the receptive fields of the kernels decrease, allowing for a more detailed neighborhood propagation of geometric priors.

\section{Conclusion}
In this paper, we proposed the tri-perspective view decomposition (TPVD), a new and novel framework for the 2D depth completion task. It decomposed the raw 3D point cloud into three 2D views to densify sparse measurements, while TPV fusion was designed to learn the 3D geometric priors via recurrent 2D-3D-2D aggregation. In view of the varying LiDAR point distributions, we introduced the distance-aware spherical convolution to refine the geometry in a compact spherical space. Moreover, we presented the geometric spatial propagation network to further improve the geometric consistency. Owing to these designs, TPVD achieves state-of-the-art performance on four benchmarks, including our newly collected dataset, TOFDC. 

\section*{Acknowledgements}
The authors thank the reviewers for their constructive comments. This work was supported by the Postgraduate Research \& Practice Innovation Program of Jiangsu Province (KYCX23\_0471), and the National Natural Science Fund of China (62361166670, 62072242, and 62376121).


{
    \small
    \bibliographystyle{ieeenat_fullname}
    \bibliography{main}

\begin{thebibliography}{62}
\providecommand{\natexlab}[1]{#1}
\providecommand{\url}[1]{\texttt{#1}}
\expandafter\ifx\csname urlstyle\endcsname\relax
  \providecommand{\doi}[1]{doi: #1}\else
  \providecommand{\doi}{doi: \begingroup \urlstyle{rm}\Url}\fi

\bibitem[Caesar et~al.(2020)Caesar, Bankiti, Lang, Vora, Liong, Xu, Krishnan, Pan, Baldan, and Beijbom]{caesar2020nuscenes}
Holger Caesar, Varun Bankiti, Alex~H Lang, Sourabh Vora, Venice~Erin Liong, Qiang Xu, Anush Krishnan, Yu Pan, Giancarlo Baldan, and Oscar Beijbom.
\newblock nuscenes: A multimodal dataset for autonomous driving.
\newblock In \emph{CVPR}, pages 11621--11631, 2020.

\bibitem[Chen et~al.(2023)Chen, Huang, Song, Deng, and Jia]{chen2023agg}
Dongyue Chen, Tingxuan Huang, Zhimin Song, Shizhuo Deng, and Tong Jia.
\newblock Agg-net: Attention guided gated-convolutional network for depth image completion.
\newblock In \emph{ICCV}, pages 8853--8862, 2023.

\bibitem[Chen et~al.(2019)Chen, Yang, Liang, and Urtasun]{chen2019learning}
Yun Chen, Bin Yang, Ming Liang, and Raquel Urtasun.
\newblock Learning joint 2d-3d representations for depth completion.
\newblock In \emph{ICCV}, pages 10023--10032, 2019.

\bibitem[Cheng et~al.(2021)Cheng, Agia, Ren, Li, and Bingbing]{cheng2021S3CNet}
Ran Cheng, Christopher Agia, Yuan Ren, Xinhai Li, and Liu Bingbing.
\newblock S3cnet: A sparse semantic scene completion network for lidar point clouds.
\newblock In \emph{CoRL}, pages 2148--2161, 2021.

\bibitem[Cheng et~al.(2018)Cheng, Wang, and Yang]{2018Learning}
Xinjing Cheng, Peng Wang, and Ruigang Yang.
\newblock Learning depth with convolutional spatial propagation network.
\newblock In \emph{ECCV}, pages 103--119, 2018.

\bibitem[Cheng et~al.(2019)Cheng, Wang, and Yang]{cspn_pami}
Xinjing Cheng, Peng Wang, and Ruigang Yang.
\newblock Learning depth with convolutional spatial propagation network.
\newblock \emph{IEEE Transactions on Pattern Analysis and Machine Intelligence}, 42\penalty0 (10):\penalty0 2361--2379, 2019.

\bibitem[Cheng et~al.(2020)Cheng, Wang, Guan, and Yang]{Cheng2020CSPN++}
Xinjing Cheng, Peng Wang, Chenye Guan, and Ruigang Yang.
\newblock Cspn++: Learning context and resource aware convolutional spatial propagation networks for depth completion.
\newblock In \emph{AAAI}, pages 10615--10622, 2020.

\bibitem[Eldesokey et~al.(2020)Eldesokey, Felsberg, and Khan]{2020Confidence}
Abdelrahman Eldesokey, Michael Felsberg, and Fahad~Shahbaz Khan.
\newblock Confidence propagation through cnns for guided sparse depth regression.
\newblock \emph{IEEE Transactions on Pattern Analysis and Machine Intelligence}, 42\penalty0 (10):\penalty0 2423--2436, 2020.

\bibitem[Gaidon et~al.(2016)Gaidon, Wang, Cabon, and Vig]{gaidon2016virtual}
Adrien Gaidon, Qiao Wang, Yohann Cabon, and Eleonora Vig.
\newblock Virtual worlds as proxy for multi-object tracking analysis.
\newblock In \emph{CVPR}, pages 4340--4349, 2016.

\bibitem[Geiger et~al.(2012)Geiger, Lenz, and Urtasun]{geiger2012we}
Andreas Geiger, Philip Lenz, and Raquel Urtasun.
\newblock Are we ready for autonomous driving? the kitti vision benchmark suite.
\newblock In \emph{CVPR}, pages 3354--3361, 2012.

\bibitem[He et~al.(2021)He, Zhu, Li, Bai, Cong, Zhang, Lin, Liu, and Zhao]{he2021towards}
Lingzhi He, Hongguang Zhu, Feng Li, Huihui Bai, Runmin Cong, Chunjie Zhang, Chunyu Lin, Meiqin Liu, and Yao Zhao.
\newblock Towards fast and accurate real-world depth super-resolution: Benchmark dataset and baseline.
\newblock In \emph{CVPR}, pages 9229--9238, 2021.

\bibitem[Hu et~al.(2021)Hu, Wang, Li, Ning, Fan, and Gong]{hu2020PENet}
Mu Hu, Shuling Wang, Bin Li, Shiyu Ning, Li Fan, and Xiaojin Gong.
\newblock Penet: Towards precise and efficient image guided depth completion.
\newblock In \emph{ICRA}, 2021.

\bibitem[Huang et~al.(2016)Huang, Sun, Liu, Sedra, and Weinberger]{huang2016deep}
Gao Huang, Yu Sun, Zhuang Liu, Daniel Sedra, and Kilian~Q Weinberger.
\newblock Deep networks with stochastic depth.
\newblock In \emph{ECCV}, pages 646--661. Springer, 2016.

\bibitem[Huang et~al.(2023)Huang, Zheng, Zhang, Zhou, and Lu]{huang2023tri}
Yuanhui Huang, Wenzhao Zheng, Yunpeng Zhang, Jie Zhou, and Jiwen Lu.
\newblock Tri-perspective view for vision-based 3d semantic occupancy prediction.
\newblock In \emph{CVPR}, pages 9223--9232, 2023.

\bibitem[Huynh et~al.(2021)Huynh, Nguyen, Matas, Rahtu, and Heikkil{\"a}]{huynh2021boosting}
Lam Huynh, Phong Nguyen, Ji{\v{r}}{\'\i} Matas, Esa Rahtu, and Janne Heikkil{\"a}.
\newblock Boosting monocular depth estimation with lightweight 3d point fusion.
\newblock In \emph{ICCV}, pages 12767--12776, 2021.

\bibitem[Imran et~al.(2021)Imran, Liu, and Morris]{imran2021depth}
Saif Imran, Xiaoming Liu, and Daniel Morris.
\newblock Depth completion with twin surface extrapolation at occlusion boundaries.
\newblock In \emph{CVPR}, pages 2583--2592, 2021.

\bibitem[Janoch et~al.(2013)Janoch, Karayev, Jia, Barron, Fritz, Saenko, and Darrell]{janoch2013category}
Allison Janoch, Sergey Karayev, Yangqing Jia, Jonathan~T Barron, Mario Fritz, Kate Saenko, and Trevor Darrell.
\newblock A category-level 3d object dataset: Putting the kinect to work.
\newblock \emph{Consumer Depth Cameras for Computer Vision: Research Topics and Applications}, pages 141--165, 2013.

\bibitem[Jaritz et~al.(2018)Jaritz, De~Charette, Wirbel, Perrotton, and Nashashibi]{2018Sparse}
Maximilian Jaritz, Raoul De~Charette, Emilie Wirbel, Xavier Perrotton, and Fawzi Nashashibi.
\newblock Sparse and dense data with cnns: Depth completion and semantic segmentation.
\newblock In \emph{3DV}, pages 52--60, 2018.

\bibitem[Kam et~al.(2022)Kam, Kim, Kim, Park, and Lee]{kam2022costdcnet}
Jaewon Kam, Jungeon Kim, Soongjin Kim, Jaesik Park, and Seungyong Lee.
\newblock Costdcnet: Cost volume based depth completion for a single rgb-d image.
\newblock In \emph{ECCV}, pages 257--274. Springer, 2022.

\bibitem[Kingma and Ba(2014)]{Kingma2014Adam}
Diederik~P Kingma and Jimmy Ba.
\newblock Adam: A method for stochastic optimization.
\newblock In \emph{Computer Ence}, 2014.

\bibitem[Ku et~al.(2018)Ku, Harakeh, and Waslander]{ku2018defense}
Jason Ku, Ali Harakeh, and Steven~L Waslander.
\newblock In defense of classical image processing: Fast depth completion on the cpu.
\newblock In \emph{CRV}, pages 16--22, 2018.

\bibitem[Lin et~al.(2022)Lin, Cheng, Zhong, Zhou, and Yang]{lin2022dynamic}
Yuankai Lin, Tao Cheng, Qi Zhong, Wending Zhou, and Hua Yang.
\newblock Dynamic spatial propagation network for depth completion.
\newblock In \emph{AAAI}, pages 1638--1646, 2022.

\bibitem[Lin et~al.(2023)Lin, Yang, Cheng, Zhou, and Yin]{lin2023dyspn}
Yuankai Lin, Hua Yang, Tao Cheng, Wending Zhou, and Zhouping Yin.
\newblock Dyspn: Learning dynamic affinity for image-guided depth completion.
\newblock \emph{IEEE Transactions on Circuits and Systems for Video Technology}, pages 1--1, 2023.

\bibitem[Liu et~al.(2023{\natexlab{a}})Liu, Kumar, Gu, Timofte, and Van~Gool]{liu2023single}
Ce Liu, Suryansh Kumar, Shuhang Gu, Radu Timofte, and Luc Van~Gool.
\newblock Single image depth prediction made better: A multivariate gaussian take.
\newblock In \emph{CVPR}, pages 17346--17356, 2023{\natexlab{a}}.

\bibitem[Liu et~al.(2021)Liu, Song, Lyu, Diao, Wang, Liu, and Zhang]{liu2021fcfr}
Lina Liu, Xibin Song, Xiaoyang Lyu, Junwei Diao, Mengmeng Wang, Yong Liu, and Liangjun Zhang.
\newblock Fcfr-net: Feature fusion based coarse-to-fine residual learning for depth completion.
\newblock In \emph{AAAI}, pages 2136--2144, 2021.

\bibitem[Liu et~al.(2023{\natexlab{b}})Liu, Song, Sun, Lyu, Li, Liu, and Zhang]{liu2023mff}
Lina Liu, Xibin Song, Jiadai Sun, Xiaoyang Lyu, Lin Li, Yong Liu, and Liangjun Zhang.
\newblock Mff-net: Towards efficient monocular depth completion with multi-modal feature fusion.
\newblock \emph{IEEE Robotics and Automation Letters}, 8\penalty0 (2):\penalty0 920--927, 2023{\natexlab{b}}.

\bibitem[Liu et~al.(2017)Liu, De~Mello, Gu, Zhong, Yang, and Kautz]{liu2017SPN}
Sifei Liu, Shalini De~Mello, Jinwei Gu, Guangyu Zhong, Ming-Hsuan Yang, and Jan Kautz.
\newblock Learning affinity via spatial propagation networks.
\newblock In \emph{NeurIPS}, 2017.

\bibitem[Liu et~al.(2022)Liu, Shao, Wang, Li, and Wang]{liu2022graphcspn}
Xin Liu, Xiaofei Shao, Bo Wang, Yali Li, and Shengjin Wang.
\newblock Graphcspn: Geometry-aware depth completion via dynamic gcns.
\newblock In \emph{ECCV}, pages 90--107. Springer, 2022.

\bibitem[Lu et~al.(2020)Lu, Barnes, Anwar, and Zheng]{2020FromLu}
Kaiyue Lu, Nick Barnes, Saeed Anwar, and Liang Zheng.
\newblock From depth what can you see? depth completion via auxiliary image reconstruction.
\newblock In \emph{CVPR}, pages 11306--11315, 2020.

\bibitem[Ma et~al.(2019)Ma, Cavalheiro, and Karaman]{ma2018self}
Fangchang Ma, Guilherme~Venturelli Cavalheiro, and Sertac Karaman.
\newblock Self-supervised sparse-to-dense: Self-supervised depth completion from lidar and monocular camera.
\newblock In \emph{ICRA}, 2019.

\bibitem[Mayer et~al.(2016)Mayer, Ilg, Hausser, Fischer, Cremers, Dosovitskiy, and Brox]{mayer2016large}
Nikolaus Mayer, Eddy Ilg, Philip Hausser, Philipp Fischer, Daniel Cremers, Alexey Dosovitskiy, and Thomas Brox.
\newblock A large dataset to train convolutional networks for disparity, optical flow, and scene flow estimation.
\newblock In \emph{CVPR}, pages 4040--4048, 2016.

\bibitem[Park et~al.(2020)Park, Joo, Hu, Liu, and Kweon]{park2020nonlocal}
Jinsun Park, Kyungdon Joo, Zhe Hu, Chi-Kuei Liu, and In~So Kweon.
\newblock Non-local spatial propagation network for depth completion.
\newblock In \emph{ECCV}, 2020.

\bibitem[Qiu et~al.(2019)Qiu, Cui, Zhang, Zhang, Liu, Zeng, and Pollefeys]{Qiu_2019_CVPR}
Jiaxiong Qiu, Zhaopeng Cui, Yinda Zhang, Xingdi Zhang, Shuaicheng Liu, Bing Zeng, and Marc Pollefeys.
\newblock Deeplidar: Deep surface normal guided depth prediction for outdoor scene from sparse lidar data and single color image.
\newblock In \emph{CVPR}, pages 3313--3322, 2019.

\bibitem[Rho et~al.(2022)Rho, Ha, and Kim]{rho2022guideformer}
Kyeongha Rho, Jinsung Ha, and Youngjung Kim.
\newblock Guideformer: Transformers for image guided depth completion.
\newblock In \emph{CVPR}, pages 6250--6259, 2022.

\bibitem[Schonberger and Frahm(2016)]{schonberger2016structure}
Johannes~L Schonberger and Jan-Michael Frahm.
\newblock Structure-from-motion revisited.
\newblock In \emph{CVPR}, pages 4104--4113, 2016.

\bibitem[Shao et~al.(2023)Shao, Pei, Chen, Wu, and Li]{shao2023nddepth}
Shuwei Shao, Zhongcai Pei, Weihai Chen, Xingming Wu, and Zhengguo Li.
\newblock Nddepth: Normal-distance assisted monocular depth estimation.
\newblock In \emph{ICCV}, pages 7931--7940, 2023.

\bibitem[Silberman et~al.(2012)Silberman, Hoiem, Kohli, and Fergus]{silberman2012indoor}
Nathan Silberman, Derek Hoiem, Pushmeet Kohli, and Rob Fergus.
\newblock Indoor segmentation and support inference from rgbd images.
\newblock In \emph{ECCV}, pages 746--760. Springer, 2012.

\bibitem[Song et~al.(2015)Song, Lichtenberg, and Xiao]{song2015sun}
Shuran Song, Samuel~P Lichtenberg, and Jianxiong Xiao.
\newblock Sun rgb-d: A rgb-d scene understanding benchmark suite.
\newblock In \emph{CVPR}, pages 567--576, 2015.

\bibitem[Tang et~al.(2020)Tang, Tian, Feng, Li, and Tan]{tang2020learning}
Jie Tang, Fei-Peng Tian, Wei Feng, Jian Li, and Ping Tan.
\newblock Learning guided convolutional network for depth completion.
\newblock \emph{IEEE Transactions on Image Processing}, 30:\penalty0 1116--1129, 2020.

\bibitem[Uhrig et~al.(2017)Uhrig, Schneider, Schneider, Franke, Brox, and Geiger]{Uhrig2017THREEDV}
Jonas Uhrig, Nick Schneider, Lukas Schneider, Uwe Franke, Thomas Brox, and Andreas Geiger.
\newblock Sparsity invariant cnns.
\newblock In \emph{3DV}, pages 11--20, 2017.

\bibitem[Van~Gansbeke et~al.(2019)Van~Gansbeke, Neven, De~Brabandere, and Van~Gool]{vangansbeke2019}
Wouter Van~Gansbeke, Davy Neven, Bert De~Brabandere, and Luc Van~Gool.
\newblock Sparse and noisy lidar completion with rgb guidance and uncertainty.
\newblock In \emph{MVA}, pages 1--6, 2019.

\bibitem[Wang et~al.(2021)Wang, Zhang, Yan, Li, Xu, Li, and Yang]{wang2021regularizing}
Kun Wang, Zhenyu Zhang, Zhiqiang Yan, Xiang Li, Baobei Xu, Jun Li, and Jian Yang.
\newblock Regularizing nighttime weirdness: Efficient self-supervised monocular depth estimation in the dark.
\newblock In \emph{ICCV}, pages 16055--16064, 2021.

\bibitem[Wang et~al.(2023)Wang, Li, Zhang, Liu, Gao, and Dai]{wang2023lrru}
Yufei Wang, Bo Li, Ge Zhang, Qi Liu, Tao Gao, and Yuchao Dai.
\newblock Lrru: Long-short range recurrent updating networks for depth completion.
\newblock In \emph{ICCV}, pages 9422--9432, 2023.

\bibitem[Wong et~al.(2020)Wong, Fei, Tsuei, and Soatto]{wong2020unsupervised}
Alex Wong, Xiaohan Fei, Stephanie Tsuei, and Stefano Soatto.
\newblock Unsupervised depth completion from visual inertial odometry.
\newblock \emph{IEEE Robotics and Automation Letters}, 5\penalty0 (2):\penalty0 1899--1906, 2020.

\bibitem[Xiao et~al.(2013)Xiao, Owens, and Torralba]{xiao2013sun3d}
Jianxiong Xiao, Andrew Owens, and Antonio Torralba.
\newblock Sun3d: A database of big spaces reconstructed using sfm and object labels.
\newblock In \emph{ICCV}, pages 1625--1632, 2013.

\bibitem[Xu et~al.(2019)Xu, Zhu, Shi, Zhang, Bao, and Li]{Xu2019Depth}
Yan Xu, Xinge Zhu, Jianping Shi, Guofeng Zhang, Hujun Bao, and Hongsheng Li.
\newblock Depth completion from sparse lidar data with depth-normal constraints.
\newblock In \emph{ICCV}, pages 2811--2820, 2019.

\bibitem[Xu et~al.(2020)Xu, Yin, and Yao]{xu2020deformable}
Zheyuan Xu, Hongche Yin, and Jian Yao.
\newblock Deformable spatial propagation networks for depth completion.
\newblock In \emph{ICIP}, pages 913--917. IEEE, 2020.

\bibitem[Yan et~al.(2022{\natexlab{a}})Yan, Li, Wang, Zhang, Li, and Yang]{yan2022multi}
Zhiqiang Yan, Xiang Li, Kun Wang, Zhenyu Zhang, Jun Li, and Jian Yang.
\newblock Multi-modal masked pre-training for monocular panoramic depth completion.
\newblock In \emph{ECCV}, pages 378--395, 2022{\natexlab{a}}.

\bibitem[Yan et~al.(2022{\natexlab{b}})Yan, Wang, Li, Zhang, Li, Li, and Yang]{yan2022learning}
Zhiqiang Yan, Kun Wang, Xiang Li, Zhenyu Zhang, Guangyu Li, Jun Li, and Jian Yang.
\newblock Learning complementary correlations for depth super-resolution with incomplete data in real world.
\newblock \emph{IEEE Transactions on Neural Networks and Learning Systems}, 2022{\natexlab{b}}.

\bibitem[Yan et~al.(2022{\natexlab{c}})Yan, Wang, Li, Zhang, Li, and Yang]{yan2022rignet}
Zhiqiang Yan, Kun Wang, Xiang Li, Zhenyu Zhang, Jun Li, and Jian Yang.
\newblock Rignet: Repetitive image guided network for depth completion.
\newblock In \emph{ECCV}, pages 214--230, 2022{\natexlab{c}}.

\bibitem[Yan et~al.(2023{\natexlab{a}})Yan, Li, Wang, Chen, Li, and Yang]{yan2023distortion}
Zhiqiang Yan, Xiang Li, Kun Wang, Shuo Chen, Jun Li, and Jian Yang.
\newblock Distortion and uncertainty aware loss for panoramic depth completion.
\newblock In \emph{ICML}, 2023{\natexlab{a}}.

\bibitem[Yan et~al.(2023{\natexlab{b}})Yan, Li, Zhang, Li, and Yang]{yan2023rignet++}
Zhiqiang Yan, Xiang Li, Zhenyu Zhang, Jun Li, and Jian Yang.
\newblock Rignet++: Efficient repetitive image guided network for depth completion.
\newblock \emph{arXiv preprint arXiv:2309.00655}, 2023{\natexlab{b}}.

\bibitem[Yan et~al.(2023{\natexlab{c}})Yan, Wang, Li, Zhang, Li, and Yang]{yan2023desnet}
Zhiqiang Yan, Kun Wang, Xiang Li, Zhenyu Zhang, Jun Li, and Jian Yang.
\newblock Desnet: Decomposed scale-consistent network for unsupervised depth completion.
\newblock In \emph{AAAI}, pages 3109--3117, 2023{\natexlab{c}}.

\bibitem[Yan et~al.(2023{\natexlab{d}})Yan, Zheng, Wang, Li, Zhang, Chen, Li, and Yang]{yan2023learnable}
Zhiqiang Yan, Yupeng Zheng, Kun Wang, Xiang Li, Zhenyu Zhang, Shuo Chen, Jun Li, and Jian Yang.
\newblock Learnable differencing center for nighttime depth perception.
\newblock \emph{arXiv preprint arXiv:2306.14538}, 2023{\natexlab{d}}.

\bibitem[Yu et~al.(2023)Yu, Sheng, Zhou, Luo, Cao, Gu, Zhang, and Shen]{yu2023aggregating}
Zhu Yu, Zehua Sheng, Zili Zhou, Lun Luo, Si-Yuan Cao, Hong Gu, Huaqi Zhang, and Hui-Liang Shen.
\newblock Aggregating feature point cloud for depth completion.
\newblock In \emph{ICCV}, pages 8732--8743, 2023.

\bibitem[Zhang et~al.(2023)Zhang, Guo, Poggi, Zhu, Huang, and Mattoccia]{zhang2023cf}
Youmin Zhang, Xianda Guo, Matteo Poggi, Zheng Zhu, Guan Huang, and Stefano Mattoccia.
\newblock Completionformer: Depth completion with convolutions and vision transformers.
\newblock In \emph{CVPR}, pages 18527--18536, 2023.

\bibitem[Zhao et~al.(2021)Zhao, Gong, Fu, and Tao]{zhao2021adaptive}
Shanshan Zhao, Mingming Gong, Huan Fu, and Dacheng Tao.
\newblock Adaptive context-aware multi-modal network for depth completion.
\newblock \emph{IEEE Transactions on Image Processing}, 2021.

\bibitem[Zhao et~al.(2022)Zhao, Zhang, Xu, Lin, and Pfister]{zhao2022discrete}
Zixiang Zhao, Jiangshe Zhang, Shuang Xu, Zudi Lin, and Hanspeter Pfister.
\newblock Discrete cosine transform network for guided depth map super-resolution.
\newblock In \emph{CVPR}, pages 5697--5707, 2022.

\bibitem[Zhao et~al.(2023)Zhao, Zhang, Gu, Tan, Xu, Zhang, Timofte, and Van~Gool]{zhao2023spherical}
Zixiang Zhao, Jiangshe Zhang, Xiang Gu, Chengli Tan, Shuang Xu, Yulun Zhang, Radu Timofte, and Luc Van~Gool.
\newblock Spherical space feature decomposition for guided depth map super-resolution.
\newblock In \emph{ICCV}, pages 12547--12558, 2023.

\bibitem[Zheng et~al.(2023)Zheng, Zhong, Li, Gao, Zheng, Jin, Wang, Zhao, Zhou, Zhang, et~al.]{zheng2023steps}
Yupeng Zheng, Chengliang Zhong, Pengfei Li, Huan-ang Gao, Yuhang Zheng, Bu Jin, Ling Wang, Hao Zhao, Guyue Zhou, Qichao Zhang, et~al.
\newblock Steps: Joint self-supervised nighttime image enhancement and depth estimation.
\newblock \emph{arXiv preprint arXiv:2302.01334}, 2023.

\bibitem[Zhou et~al.(2023)Zhou, Yan, Liao, Lin, Huang, Zhao, Cui, and Li]{zhou2023bev}
Wending Zhou, Xu Yan, Yinghong Liao, Yuankai Lin, Jin Huang, Gangming Zhao, Shuguang Cui, and Zhen Li.
\newblock Bev@ dc: Bird's-eye view assisted training for depth completion.
\newblock In \emph{CVPR}, pages 9233--9242, 2023.

\bibitem[Zuo et~al.(2023)Zuo, Zheng, Huang, Zhou, and Lu]{zuo2023pointocc}
Sicheng Zuo, Wenzhao Zheng, Yuanhui Huang, Jie Zhou, and Jiwen Lu.
\newblock Pointocc: Cylindrical tri-perspective view for point-based 3d semantic occupancy prediction.
\newblock \emph{arXiv preprint arXiv:2308.16896}, 2023.

\end{thebibliography}
}

\clearpage
\setcounter{page}{1}
\maketitlesupplementary

 \begin{figure}[t]
  \centering
  \includegraphics[width=1\columnwidth]{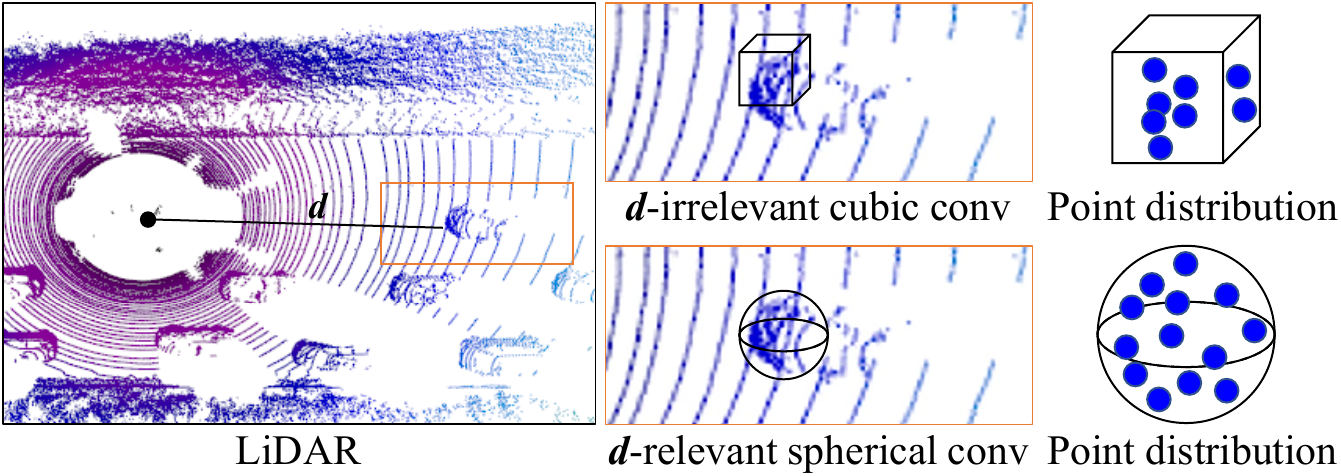}\\
  \caption{Comparison of the common 3D cubic convolution and our proposed distance-aware spherical convolution. }\label{fig_pd}
\end{figure}

\begin{table}[t]
\centering
\renewcommand\arraystretch{1.5}
\resizebox{0.478\textwidth}{!}{
\begin{tabular}{ll}
\toprule
\multicolumn{2}{l}{For one pixel $p$ in the valid pixel set $\mathbb{P}$: } \\ \midrule
-- REL               & $\frac{1}{|\mathbb{P}|}\sum\limits{{{\left| \mathbf y_{p}-{\mathbf x}_{p} \right|/{\mathbf y_{p}}}}}$  \\ 
-- MAE               & $\frac{1}{|\mathbb{P}|}\sum\limits{{{\left| \mathbf y_{p}-{\mathbf x}_{p} \right|}}}$  \\ 
-- iMAE              & $\frac{1}{|\mathbb{P}|}\sum\limits{{{\left| 1/{\mathbf y_{p}}-1/{{\mathbf x}_{p}} \right|}}}$  \\ 
-- RMSE              & $\sqrt{\frac{1}{|\mathbb{P}|}\sum\limits{{{\left( \mathbf y_{p}-{\mathbf x}_{p} \right)}}}^2}$  \\  
-- iRMSE             & $\sqrt{\frac{1}{|\mathbb{P}|}\sum\limits{{{\left( 1/{\mathbf y_{p}}-1/{{\mathbf x}_{p}} \right)}}}^2}$  \\  
-- RMSELog           & $\sqrt{\frac{1}{|\mathbb{P}|}\sum{{{\left(  {\log {\mathbf y}}-\log {\mathbf x} \right)}^{2}}}}$  \\  
-- ${\delta }_{i}$   & $\frac{|\mathbb{S}|}{|\mathbb{P}|}, \ \mathbb{S}: \max \left( {{\mathbf y_p}/{\mathbf x_p},{\mathbf x_p}/{\mathbf y_p}} \right)<{1.25}^{i}$  \\ 
\bottomrule
\end{tabular}
}
\caption{Definition of the seven metrics used in the main text.}\label{tab_metric}
\end{table}

\section{Distance-Aware Spherical Convolution}
\cref{fig_pd} illustrates the comparison of our distance-aware spherical convolution (DASC) and the 3D convolution. We observe that the $d$-relevant DASC involves a higher number of valid points with more balanced distribution.

\section{Metric}
On KITTI benchmark, we employ RMSE, MAE, iRMSE, and iMAE for evaluation \cite{park2020nonlocal,zhou2023bev,wang2023lrru,yan2022rignet}. On NYUv2, TOFDC, and SUN RGBD datasets, RMSE, REL, and ${{\delta }_{i}}$ ($i=1, 2, 3$) are selected for testing \cite{Qiu_2019_CVPR,tang2020learning,zhang2023cf,yu2023aggregating}. 

For simplicity, let $\mathbf{x}$ and $\mathbf{y}$ denote the predicted depth and ground truth depth, respectively. \cref{tab_metric} defines the metrics.

\section{Loss Function}
The total loss function $L_{total}$ consists of three terms, \textit{i.e.}, the front-view $L_f$, top-view $L_t$, and side-view $L_s$. The ground truths of the front, top, and side views are obtained by projecting the annotated point clouds. Following \cite{park2020nonlocal,lin2023dyspn,zhou2023bev}, we adopt $L_1$ and $L_2$ joint loss functions to denote $L_f$, $L_t$, and $L_s$, \textit{i.e.}, $L_f/L_t/L_s=L_1+L_2$. As a result, the total loss function $L_{total}$ is defined as: 
\begin{equation}\label{loss_function}
    L_{total}=L_f + \alpha L_t + \beta L_s,
\end{equation}
where $\alpha$ and $\beta$ are conducted to balance the three terms. Empirically, we set $\alpha$ and $\beta$ to 0.6 and 0.2, respectively. 

\section{Implementation Detail}
We implement TPVD on Pytorch with four 3090 GPUs. We train it for 50 epochs with Adam \cite{Kingma2014Adam} optimizer. The initial learning rate is $5\times 10^{-4}$ for the first 30 epochs and is reduced to half for every 10 epochs. Following \cite{lin2023dyspn,wang2023lrru}, the stochastic depth strategy \cite{huang2016deep} is used for better training. Also, we employ color jitter and random horizontal flip for data augmentation. The batch size is 3 for each GPU.

\begin{table*}[t]
\centering
\renewcommand\arraystretch{1.1}
\resizebox{0.87\textwidth}{!}{
\begin{tabular}{l|cc|c|c|ccc|c}
\toprule
Dataset     & Outdoor & Indoor  & Sensor  & Edge Device  & Train  & Test  & Resolution  & Real-world  \\ 
\midrule
KITTI \cite{Uhrig2017THREEDV}    & \checkmark & $\times$    & LiDAR       & $\times$  & 86,898  & 1,000  & $1216\ast 352$  & \checkmark  \\
NYUv2 \cite{silberman2012indoor} &  $\times$  & \checkmark  & Kinect TOF  & $\times$  & 47,584  & 654    & $304\ast 228$  & $\times$  \\
TOFDC                            & \checkmark & \checkmark  & Phone TOF   & \checkmark  & 10,000  & 560    & $512\ast 384$  & \checkmark  \\
\bottomrule
\end{tabular}
}
\vspace{-5pt}
\caption{Dataset comparison. Note that these characteristics are calculated according to the \textbf{depth completion task}.}\label{tab_dataset_comparison}
\vspace{-5pt}
\end{table*}

\section{TOFDC}

\subsection{Motivation}
For depth completion task, the commonly used datasets are KITTI \cite{Uhrig2017THREEDV} and NYUv2 \cite{silberman2012indoor}. \cref{tab_dataset_comparison} lists the detailed characteristics. 
KITTI uses LiDAR to collect outdoor scenes, while NYUv2 employs Kinect with time-of-flight (TOF) to capture indoor scenes. However, both LiDAR and Kinect are bulky and inconvenient, especially for ordinary consumers in daily life. Recently, TOF depth sensors have become more common on edge devices (\textit{e.g.}, mobile phones), as depth information is vital for human-computer interaction, such as virtual reality and augmented reality. Therefore, it is important and worthwhile to create a new depth completion dataset on consumer-level edge devices.

\subsection{Data Collection}
\noindent \textbf{Acquisition System.} As illustrated in \cref{collection_raw_data} (left), 
the acquisition system consists of the Huawei P30 Pro and Helios, which capture color image and raw depth, and ground truth depth, respectively. The color camera of P30 produces $3648\times 2736$ color images using a 40 megapixel Quad Bayer RYYB sensor, while the TOF camera outputs $240\times 180$ raw depth maps. The industrial-level Helios TOF camera generates higher-resolution depth. Their depth acquisition principle is the same, ensuring consistent depth values. 

 \begin{figure}[t]
  \centering
  \includegraphics[width=1\columnwidth]{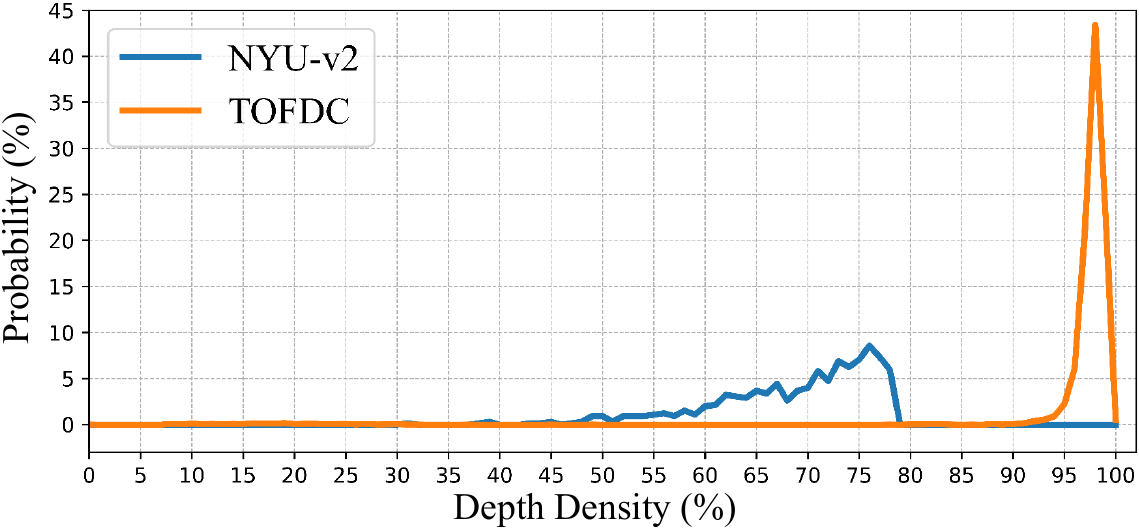}\\
  \caption{Density-probability comparison of raw depth maps.}\label{tofdc_statistic_density_prob}
\end{figure}

\noindent \textbf{Data Processing.} We calibrate the RGB-D system of the P30 with the Helios TOF camera. We align them on the $640\times 480$ color image coordinate using the intrinsic and extrinsic parameters. The color images and Helios depth maps are cropped to $512\times 384$, while the P30 depth maps to $192\times 144$. Then we conduct nearest interpolation to upsample the P30 depth maps to $512\times 384$. For the Helios depth maps, there still exist some depth holes caused by environment and object materials (\textit{e.g.}, transparent glass). We use the colorization technique (Levin \textit{et. al}) to fill the holes. \cref{collection_raw_data} (right) shows the visual result. 

\cref{tofdc_statistic_density_prob} provides the corresponding statistical support. It reveals that the depth density of NYUv2 varies mainly from 60\% to 80\%, whereas that of TOFDC is highly concentrated between 95\% and 100\%.

\begin{figure}[!ht]
  \centering
  \includegraphics[width=1\columnwidth]{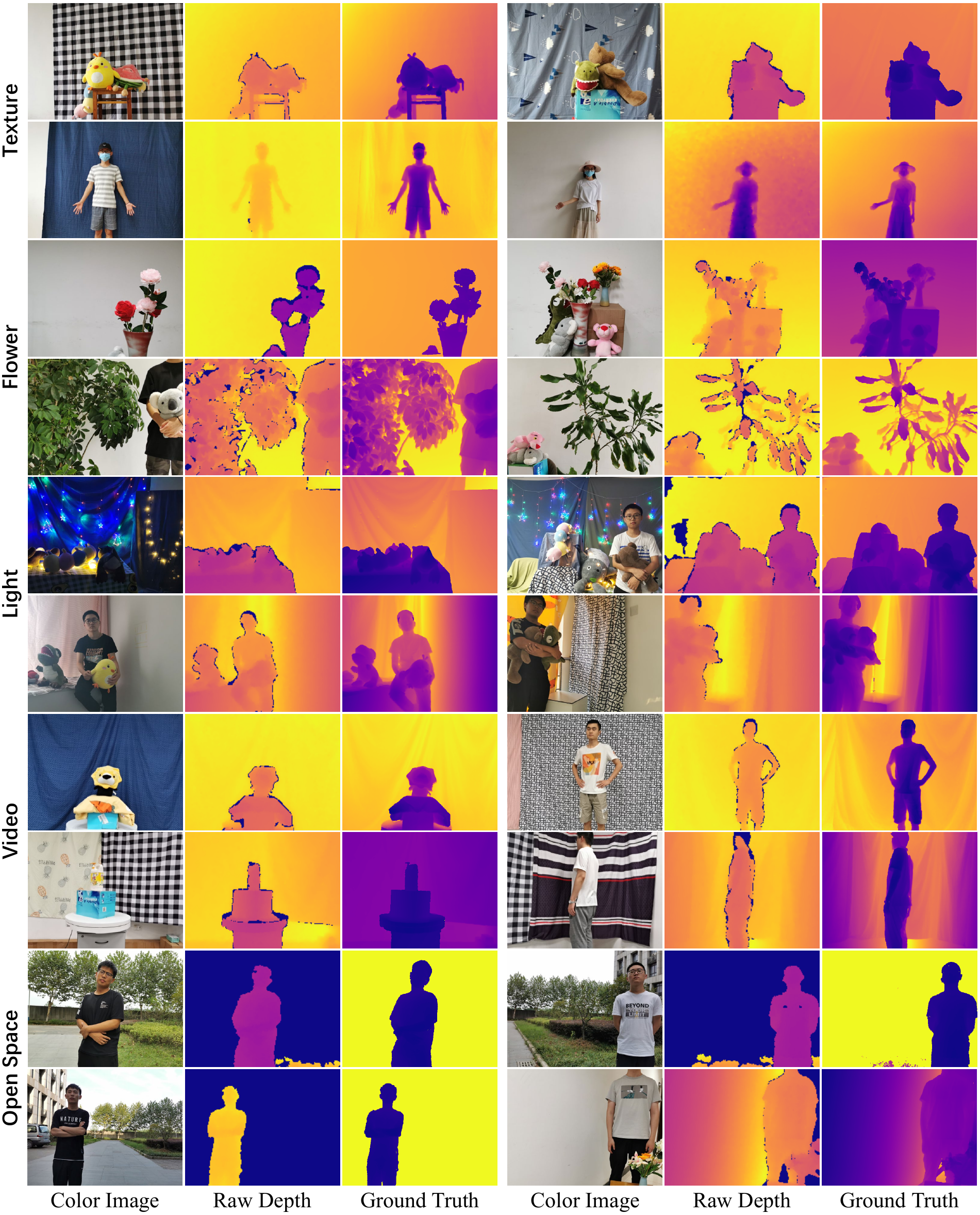}\\
  \vspace{-5pt}
  \caption{TOFDC examples in different scenarios.}\label{tofdc_examples}
\end{figure}

As reported in \cref{tab_dataset_comparison}, we collect the new depth completion dataset TOFDC. It consists of indoor and outdoor scenes, including texture, flower, light, video, and open space in \cref{tofdc_examples}. For the depth completion task, we take the raw depth captured by the P30 TOF lens as input, which is different from NYUv2 where the input depth is sampled from the ground truths. 

\subsection{Cross-Dataset Evaluation}
To validate the generalization on indoor scenes~\cite{yu2023aggregating}, we train TPVD on NYUv2 and test it on SUN RGBD. Comparing Tab.~\ref{tab_sunrgbd}-Kinect with Tab.~\ref{tab_nyuv2}, the errors of all methods increase and the accuracy decreases due to different RGB-D sensors. When comparing Tab.~\ref{tab_sunrgbd}-Xtion with Tab.~\ref{tab_nyuv2}, since the data is from different Xtion devices, 
we discover that the performance drops by large margins. However, Tab.~\ref{tab_sunrgbd} reports that our TPVD still 
achieves the lowest errors and the highest accuracy under Kinect V1 and Xtion splits. For example, under Xtion split, the RMSE of TPVD is 9 mm superior to those of the second best NLSPN~\cite{park2020nonlocal} and PointDC~\cite{yu2023aggregating}. These facts evidence the powerful cross-dataset generalization ability of our TPVD.

\begin{table}[t]
\centering
\renewcommand\arraystretch{1.1}
\resizebox{0.478\textwidth}{!}{
\begin{tabular}{l|ccccc}
\toprule
Method          & RMSE (m) $\downarrow$      & REL $\downarrow$     &${\delta }_{1}$  $\uparrow$ & ${\delta }_{{2}}$ $\uparrow$ & ${\delta }_{{3}}$  $\uparrow$ \\ 
\midrule 
\multicolumn{6}{c}{Collected by Kinect V1} \\ 
\midrule 
CSPN \cite{2018Learning}          & 0.729    & 0.504   & 69.1  & 77.8  & 84.0  \\
NLSPN \cite{park2020nonlocal}     & 0.093    & 0.020   & 98.9  & 99.6  & 99.7  \\ 
CostDCNet \cite{kam2022costdcnet} & 0.119    & 0.033   & 98.1  & 99.6  & 99.7  \\ 
GraphCSPN \cite{liu2022graphcspn} & 0.094    & \textcolor{blue}{0.023}   & 98.8  & 99.6  & 99.7  \\ 
PointDC \cite{yu2023aggregating}  & \textcolor{blue}{0.092}    & \textcolor{blue}{0.023}   & 98.9  & 99.6  & 99.8  \\
\textbf{TPVD (ours) }             & \textbf{0.087}    & \textbf{0.022}   & \textbf{99.1}  & \textbf{99.7}  & \textbf{99.8}  \\
\midrule
\multicolumn{6}{c}{Collected by Xtion} \\ 
\midrule 
CSPN \cite{2018Learning}          & 0.490    & 0.179   & 84.5  & 91.5  & 95.1  \\
NLSPN \cite{park2020nonlocal}     & \textcolor{blue}{0.128}    & 0.015   & 99.0  & 99.7  & 99.9  \\ 
CostDCNet \cite{kam2022costdcnet} & 0.207    & 0.028   & 97.8  & 99.1  & 99.5  \\ 
GraphCSPN \cite{liu2022graphcspn} & 0.131    & 0.017   & 99.0  & 99.7  & 99.9  \\ 
PointDC \cite{yu2023aggregating}  & \textcolor{blue}{0.128}    & \textcolor{blue}{0.016}   & 99.1  & 99.7  & 99.9  \\
\textbf{TPVD (ours) }             & \textbf{0.119}    & \textbf{0.014}   & \textbf{99.3}  & \textbf{99.8}  & \textbf{99.9}  \\
\bottomrule
\end{tabular}
}
\vspace{-6pt}
\caption{Cross-dataset evaluation on SUN RGBD benchmark.}\label{tab_sunrgbd}
\vspace{-4pt}
\end{table}

\end{document}